%% file: main.tex
\documentclass[lettersize,journal]{IEEEtran}
\usepackage{amsmath,amsfonts}
\usepackage{algorithmic}
\usepackage{algorithm}
\usepackage{array}
\usepackage[caption=false,font=normalsize,labelfont=sf,textfont=sf]{subfig}
\usepackage{textcomp}
\usepackage{stfloats}
\usepackage{url}
\usepackage{verbatim}
\usepackage{graphicx}
\usepackage{cite}
\usepackage{booktabs}
\usepackage{multirow}
\usepackage{xcolor}
\usepackage{amssymb,bm}
\usepackage[algo2e,ruled,linesnumbered]{algorithm2e}
\usepackage{placeins}
\usepackage[hidelinks]{hyperref}
\providecommand{\IEEEPARstart}[2]{#1#2}
\renewenvironment{IEEEkeywords}{\par\smallskip\noindent\textbf{Keywords: }\ignorespaces}{\par\smallskip}

\newcommand{\gsp}{GSAP}

\newcommand{\edit}[1]{#1}
\newcommand{\accrev}[1]{#1}
\newcommand{\newrev}[1]{#1}

\newcommand{\absrel}{AbsRel}
\newcommand{\rmse}{RMSE}

\input{table}

\input{figure}

\title{\edit{GeoBridge: Evolving Transferable Decoder Topologies for Geometric Vision}}
\author{Rui Wang, Haoyu Zhang, and Ran Cheng%
\thanks{Rui Wang and Haoyu Zhang are with the School of Information Science and Technology, Hangzhou Normal University, Hangzhou, P.R. China (e-mail: kujiyunmili@gmail.com; haoyu.zhang@hznu.edu.cn).}%
\thanks{Ran Cheng is with \accrev{the Department of Data Science and Artificial Intelligence, The Hong Kong Polytechnic University}, Hong Kong SAR, China (e-mail: ranchengcn@gmail.com).}%
}
\begin{document}
\maketitle
\begin{abstract}
\edit{Depth estimation and Gaussian reconstruction share a need for multi-scale feature processing, yet use different prediction interfaces and training objectives. Whether a decoder's internal computation can be optimized once and reused across these systems remains a structural design question. \accrev{Search over primitive configurations alone leaves the organization of transformations fixed.} To address this gap, we propose GeoBridge, an evolutionary framework for transferable decoder topologies. First, GeoBridge makes stage-internal topology searchable through residual, sequential, and parallel composition of configurable primitives. Shape-preserving composition retains the stage interfaces of a fixed multi-scale scaffold. Second, a functional-tree encoding supports stage-aware crossover and hierarchical mutation, while a gradient sensitivity-alignment proxy evaluates candidates without training. Multi-objective evolution balances this proxy with computation and parameter cost. Third, selected topologies are instantiated and trained in target systems with task-specific prediction interfaces. A single source search takes approximately 12 GPU hours. GeoBridgeNet-A3 achieves the best depth accuracy among four adapted decoder baselines under a common encoder and training protocol on NYU Depth V2 and KITTI. The same topology reaches 27.11~dB PSNR in Gaussian novel-view synthesis on RealEstate10K, and its refined representations support open-vocabulary segmentation on LERF. These results support decoder topology as a reusable design object across the evaluated geometric vision systems.}
\edit{Code and pretrained weights are available at \url{https://github.com/EMI-Group/GeoBridge-NAS}.}
\end{abstract}


\section{Introduction}
\label{sec:introduction}

\edit{\IEEEPARstart{D}{ecoders} connect visual features to geometric predictions through multi-scale processing and spatial detail recovery. In monocular depth estimation (MDE), they map image features to dense depth~\cite{eigen2014depth,bhat2021adabins,yuan2022newcrfs}. In feed-forward Gaussian reconstruction, decoded features contribute to depth inference and the prediction of a renderable scene representation~\cite{charatan2024pixelsplat,chen2024mvsplat,xu2025depthsplat}. These systems have different prediction interfaces and training objectives, yet share a need to organize feature transformations across spatial scales. This common role motivates a structural question: can a decoder's internal computation be optimized in one geometric setting and remain useful in another?}

\edit{Answering this question requires treating the organization of decoder computation as a design variable. In a decoder stage with a prescribed composition, changing a primitive's depth, scan pattern, or fusion rule leaves the surrounding serial, parallel, and residual arrangement fixed. These arrangements determine how transformations accumulate along a path and how branch outputs are aggregated. Neural architecture search (NAS) has explored operation connectivity, multi-scale fusion, and hierarchical motifs~\cite{liu2019darts,ghiasi2019nasfpn,liu2018hierarchical}. For decoder reuse across geometric systems, the specific design problem is to expose stage-internal composition to search while retaining the interfaces needed by the surrounding network.}

\edit{We investigate the hypothesis that decoder topology can serve as a reusable design object across geometric prediction systems. Here, topology specifies the residual, sequential, and parallel organization of transformations within each decoder stage. \accrev{Transfer retains this organization and the primitive configurations; each target system trains its own weights and prediction interface.} A fixed multi-scale scaffold supplies compatible stage boundaries, and variable internal topologies provide alternative ways to process the features at those boundaries. This separation permits one source search followed by target training under different prediction objectives.}

\edit{To realize this design, we propose GeoBridge, an evolutionary framework that searches decoder topologies in a depth-estimation scaffold and reuses the selected structures in target geometric systems. A bounded composition grammar constructs each stage from shape-preserving primitives through residual, sequential, and parallel operators. Functional trees encode these compositions, so stage-aware crossover and hierarchical mutation can vary complete stages, subtrees, and primitive configurations. A gradient sensitivity-alignment proxy (GSAP) evaluates independently initialized candidates under a common teacher-derived regression stimulus. NSGA-II~\cite{deb2002nsga2} balances this training-free score with computation and parameter cost to obtain a set of decoder structures for subsequent reuse.}

\edit{Experiments test whether the structures selected in this source setting retain their utility after target training. One search on TartanAir~\cite{wang2020tartanair} produces the GeoBridgeNet family in approximately 12 GPU hours. GeoBridgeNet-A3 achieves the best depth accuracy among four recent decoder baselines under a common encoder and training protocol on NYU Depth V2 and KITTI. The same topology reaches 27.11~dB PSNR after integration into a Gaussian novel-view synthesis (NVS) system on RealEstate10K. Its Gaussian outputs also support open-vocabulary segmentation on LERF after scene refinement and semantic optimization~\cite{kerr2023lerf,kerbl2023gaussians,zhu2025cos3d}. Controlled studies characterize the decoder designs, the structural coverage of evolutionary variation, and the ranking quality of GSAP.}

The main contributions of this work are summarized as follows:
\begin{itemize}
    \item \edit{We formulate decoder design as evolution of stage-internal computation topologies within a fixed multi-scale scaffold. A shape-preserving composition grammar opens residual, sequential, and parallel organization to search alongside primitive configuration.}
    \item \edit{We develop an evolutionary search over these compositions with stage-aware crossover, hierarchical mutation, and GSAP evaluation. This combination supports independent candidate instantiation and multi-objective selection without candidate training.}
    \item \edit{We demonstrate reuse of the selected decoder structures across real-world depth estimation and Gaussian novel-view synthesis through target-specific training. The evaluation connects a single source search to prediction systems with different output representations and further examines the semantic utility of their reconstructed scenes.}
\end{itemize}

\edit{Section~\ref{sec:related_work} reviews decoder design, architecture transfer, structural encoding, and candidate evaluation. Section~\ref{sec:method} presents the topology space, evolutionary search, and target instantiation. Section~\ref{sec:experiments} evaluates the resulting structures and their reuse, followed by conclusions in Section~\ref{sec:conclusion}.}

\section{Related Work}
\label{sec:related_work}

\edit{The design of transferable decoder topologies connects architecture search with structural reuse. We first review how NAS exposes visual computation to optimization and how architectures are transferred between prediction tasks. We then examine the representations and evaluation methods that support evolutionary search over composed structures.}

\subsection{\edit{Architecture Search for Visual Decoding}}
\label{sec:rw_nas}

\edit{NAS combines an architecture space, a search strategy, and a performance estimator to automate network design. Early studies focused on image classification and established reinforcement learning~\cite{zoph2018nasnet}, gradient-based optimization~\cite{liu2019darts}, and evolutionary search~\cite{lu2019nsganet} as major search strategies. Evolutionary NAS directly manipulates discrete operations and connections through selection, crossover, and mutation. Its population-based formulation also supports the joint optimization of prediction quality and computational cost.}

\edit{Dense prediction has motivated architecture spaces tailored to multi-scale processing. NAS-FPN and FNA++ search feature fusion and network structures for object detection~\cite{ghiasi2019nasfpn,fang2021fnapp}. Auto-DeepLab searches multi-resolution structures for semantic segmentation~\cite{liu2019autodeeplab}, while SCTNet-NAS incorporates deployment efficiency~\cite{liu2025sctnetnas}. MDE-oriented methods further specialize search to depth prediction~\cite{huynh2022lidnas,yu2025mdeevonas,zhang2025ptfevomde}.} \accrev{Task-specific search spaces also address feature extraction and fusion in other vision problems. TREFE searches the spatial and sequential feature extractors of text recognition networks under a latency constraint~\cite{zhang2023trefe}, and NAS-PED searches hybrid convolution--transformer backbones for pedestrian detection in crowded scenes~\cite{tang2025nasped}. SPOSGait applies single-path one-shot search to gait recognition in unconstrained environments~\cite{guo2025sposgait}, while DCNAS searches cross-modality feature extraction and fusion structures for guided image super-resolution~\cite{zhong2025dcnas}.} \edit{These approaches establish topology and operation choice as NAS design variables. GeoBridge investigates their combination within decoder stages whose external multi-scale connections remain fixed across the search space.}

\subsection{Architecture Transfer}
\label{sec:rw_transfer}

\edit{Architecture transfer reuses design knowledge across datasets or prediction tasks. NAT adapts a pretrained supernet to target datasets~\cite{lu2021nat}, while NASNet transfers classification-oriented cells to object-detection backbones~\cite{zoph2018nasnet}. FNA++ combines parameter remapping with target-side architecture adaptation for dense prediction~\cite{fang2021fnapp}. These methods illustrate different transfer units, from selected network configurations to reusable cells and adapted networks.}

\edit{Cross-task studies examine which architectural choices remain useful when the prediction objective changes. TransNAS-Bench-101 evaluates architectures across multiple visual tasks~\cite{duan2021transnas}, and PTF-EvoMDE transfers an encoder searched for depth estimation to classification, detection, and segmentation~\cite{zhang2025ptfevomde}. LAPT transfers architectural design principles~\cite{zhou2025lapt}. GeoBridge studies the stage-internal topology of a decoder as the transfer unit between dense depth prediction and Gaussian reconstruction. Each target system retains the selected composition and primitive configurations and trains the instantiated network for its own objective.}

\subsection{Structural Encoding in EvoNAS}
\label{sec:rw_encoding}

\edit{Architecture encoding determines the units on which evolutionary variation operates~\cite{white2020encodings}. Direct encodings specify operations and connections explicitly. EvoCNN, for example, uses variable-length chromosomes to represent layer sequences~\cite{sun2019evolving}. \accrev{NASNet searches reusable normal and reduction cells and stacks them into a full network~\cite{zoph2018nasnet}. Mainstream gradient-based NAS methods, including DARTS and P-DARTS, search over this cell-based encoding~\cite{liu2019darts,chen2021pdarts}.} Such encodings expose local architectural choices to crossover and mutation, while the decoding procedure determines how those choices form a valid network.}

\accrev{Early work on evolving artificial neural networks also used indirect encodings, in which the genome describes a construction process instead of listing every unit and connection. Kitano encodes a connectivity matrix through graph-rewriting rules~\cite{kitano1990graph}, and cellular encoding grows a network from a single cell by a program of division and modification instructions~\cite{gruau1994cellular}. NEAT instead evolves explicit node and connection genes, with historical markings that align genomes during crossover~\cite{stanley2002neat}, and HyperNEAT generates connection weights from a function of neuron coordinates~\cite{stanley2009hyperneat}.}

\edit{\accrev{More recent encodings} expose larger structural units through composition. SLGE constructs architectures with grammar-based transformations~\cite{broni2020slge}, hierarchical NAS assembles motifs from lower-level operations~\cite{liu2018hierarchical}, and tree-based genetic programming has been applied to encoder--decoder networks~\cite{fuentes2024tree}. GeoBridge adopts function composition within a fixed multi-scale decoder scaffold. Its residual, sequential, and parallel nodes preserve the stage interface, so both a complete tree and any subtree define executable feature transformations. This organization supports structural variation at several levels without changing the connections between decoder stages.}

\subsection{Efficient EvoNAS}
\label{sec:related_evaluation}

\edit{Candidate evaluation is a major cost in population-based NAS. Multi-fidelity evaluation reduces the resources allocated to candidate training~\cite{yang2022multifidelity}. Weight sharing reduces this cost by evaluating candidates within a trained supernet. SPOS uses uniformly sampled single-path subnetworks~\cite{guo2020spos}, while Once-for-All supports elastic depth, width, kernel size, and input resolution~\cite{cai2020ofa}.} \accrev{BossNAS divides the supernet into blocks trained by self-supervised ensemble bootstrapping, which improves architecture ranking in hybrid convolution--transformer spaces~\cite{li2025bossnas}. WIGEvoNAS uses weight inheritance to shorten candidate training in each generation of evolutionary search~\cite{yan2026wigevonas}.} \edit{Applying these strategies to composed decoder trees requires a supernet that specifies how parameters are shared across different structural positions. GeoBridge instead evaluates independently instantiated candidates and uses a training-free proxy to control their evaluation cost.}

\edit{Surrogate-assisted EvoNAS predicts candidate quality from previously evaluated architectures. Sun et al.~\cite{sun2020e2epp} use an offline random-forest predictor, and Ma et al.~\cite{ma2024paretorank} learn Pareto-wise dominance relationships for multi-objective search.} \accrev{CFMOGP-NAS combines the forecasts of multiple regression models within multi-objective genetic programming~\cite{cao2026cfmogp}, and Li et al.~\cite{li2026fuzzynas} replace performance regression with a classifier when evolving fuzzy neural networks for defect recognition.} \edit{These approaches require performance observations to train the estimator. Parameter-transfer methods such as PTF-EvoMDE reduce candidate-specific pretraining through hierarchical parameter mapping~\cite{zhang2025ptfevomde}. Training-free proxies offer a complementary route through responses measured at initialization, including connection sensitivity, gradient flow, activation statistics, and combinations of such criteria~\cite{lee2019snip,wang2020grasp,mellor2021naswot,abdelfattah2021zerocost,lee2024aznas}. GSAP combines a parameter-scaled sensitivity statistic with gradient--parameter alignment under the regression stimulus used in the source search.}

\section{Proposed Method}
\label{sec:method}

\edit{GeoBridge evolves the internal topology of a decoder while preserving the stage interfaces through which the decoder connects to a geometric prediction system. Fig.~\ref{fig:generality} summarizes the source search and target reuse. We first define the source search and its multi-scale scaffold, then describe the composition space and training-free evaluator. The evolutionary procedure selects structures from this space, and target-specific training tests their reuse in depth estimation and Gaussian reconstruction.}

\FigureGenerality

\subsection{\edit{Design Object and Source Scaffold}}
\label{sec:problem}

\edit{The source search jointly varies stage topology and primitive configuration under a fixed decoder scaffold. Let $\mathcal{D}_{\mathrm{src}}$ denote the source dataset and $\Omega_{\alpha}$ the composition space defined in Section~\ref{sec:encoding}. An architecture $\alpha\in\Omega_{\alpha}$ specifies both the composition of each stage and the configuration of its primitive transformations. With $\bm{w}^{\ast}$ denoting parameters optimized on the source task, the training-based reference problem is}

\begin{equation}
\begin{aligned}
\operatorname*{minimize}_{\alpha \in \Omega_{\alpha}}
\quad &
\left\{
f_1(\alpha;\bm{w}^{\ast}),
C_{\mathrm{MAC}}(\alpha),
C_{\mathrm{Param}}(\alpha)
\right\}, \\
\text{subject to}\quad &
\bm{w}^{\ast}
\in
\operatorname*{arg\,min}_{\bm{w}\in\Omega_w}
\mathcal{L}(\bm{w};\alpha).
\end{aligned}
\label{eq:bridge_search}
\end{equation}

\edit{Here, $f_1(\alpha;\bm{w}^{\ast})$ measures depth prediction error after training, while $C_{\mathrm{MAC}}(\alpha)$ and $C_{\mathrm{Param}}(\alpha)$ measure decoder computation and parameter cost. These objectives describe the trade-off between predictive quality and resource use. GeoBridge maintains a non-dominated set, denoted by $\mathcal{Q}_{\mathrm{src}}$, from which decoder structures with different cost profiles are selected.}

\edit{The selected architecture is the object reused across target systems. For $\alpha^{\star}\in\mathcal{Q}_{\mathrm{src}}$, the stage compositions and primitive configurations remain fixed, while each target system optimizes its network weights and prediction interface. Thus, source search chooses the computation structure, and target training determines its parameters for each prediction objective.}

\edit{Direct evaluation of Eq.~\eqref{eq:bridge_search} requires training every candidate decoder. To reduce this cost, GeoBridge uses $J_{\mathrm{GSAP}}$ from Section~\ref{sec:GSAP_method} as the quality objective during evolution. The resulting search minimizes $\{J_{\mathrm{GSAP}},C_{\mathrm{MAC}},C_{\mathrm{Param}}\}$ over independently instantiated candidates. Section~\ref{sec:GSAP_analysis} evaluates the association between the proxy and retrained depth accuracy.}

\edit{All candidate decoders share the multi-scale scaffold shown in Fig.~\ref{fig:overall}. DINOv2 ViT-S~\cite{oquab2024dinov2} supplies intermediate features from blocks B2, B5, B8, and B11, which are projected to scales of $1/4$, $1/8$, $1/16$, and $1/32$. Four Tree-Mamba stages process these features along a coarse-to-fine pathway. TartanAir images and cached features from a frozen Depth Anything V2 teacher provide the source inputs and auxiliary regression targets used for candidate evaluation.}

\edit{The computation inside each Tree-Mamba stage is the searchable part of the scaffold. A projection adapts the encoder feature, which is fused with the preceding stage's upsampled output at all but the coarsest scale. The stage then transforms this fused feature through its internal composition, and the finest-scale output is mapped to a prediction by the source head. The projections, cross-scale connections, upsampling operators, and head remain fixed during search. Consequently, candidates differ in how each stage processes its inputs while sharing the same information flow between stages.}

\FigureOverall

\subsection{\edit{Compositional Decoder Topologies}}
\label{sec:encoding}

\edit{The topology space exposes the organization of transformations within each decoder stage. An architecture is represented by four functional trees, $\alpha=\{T_{32},T_{16},T_{8},T_{4}\}$. Leaf nodes instantiate configurable Mamba primitives, and internal nodes define residual, sequential, or parallel composition. These choices vary processing depth, branching, and residual paths while preserving the stage interface. Fig.~\ref{fig:search_space} illustrates the mapping from the tree representation to executable computation.}

\subsubsection{Stage-Wise Tree Structure}

The four functional trees are aligned with decoder resolutions of $1/32$, $1/16$, $1/8$, and $1/4$, denoted by $T_{32}$, $T_{16}$, $T_{8}$, and $T_{4}$, respectively. Each tree receives the fused feature of the corresponding Tree-Mamba stage and returns a representation with the same external interface for subsequent coarse-to-fine decoding. In this way, the fixed multi-scale scaffold determines how information flows across stages, whereas the functional tree determines how features are transformed within each stage.

\edit{Stage-specific bounds control the structural capacity of the decoder. The coarse stages $T_{32}$ and $T_{16}$ permit at most two levels of internal composition, while $T_{8}$ and $T_{4}$ permit at most one. With the child order of PAR treated as equivalent, the four one-level structures are OP, RES(OP), SEQ(OP,OP), and PAR(OP,OP). Each coarse stage therefore admits $1+4+4^2+4(4+1)/2=31$ structures, which gives $31^2\times4^2=15{,}376$ four-stage topologies. Including the primitive domains in Table~\ref{tab:search_space} gives approximately $2.5\times10^{26}$ configurations. These bounds allocate more compositional capacity to coarse stages, where spatial processing is less expensive.}

\subsubsection{Composition Nodes}

The internal nodes define how primitive or composed transformations are organized into higher-level functions. Let $\mathcal{F}$ and $\mathcal{G}$ denote two shape-preserving child transformations with input $\bm{x}$. The three composition operators are defined as

\begin{equation}
\operatorname{RES}(\bm{x})
=
\bm{x}+\mathcal{F}(\bm{x}),
\end{equation}

\begin{equation}
\operatorname{SEQ}(\bm{x})
=
\mathcal{G}\bigl(\mathcal{F}(\bm{x})\bigr),
\end{equation}

and

\begin{equation}
\operatorname{PAR}(\bm{x})
=
\frac{1}{2}
\left[
\mathcal{F}(\bm{x})+\mathcal{G}(\bm{x})
\right].
\end{equation}

\edit{The composition rules control distinct aspects of stage topology. RES adds an identity path around a transformation, SEQ places transformations along a longer processing path, and PAR aggregates two branches that receive the same input. Each composition returns a function with the same input and output shape as its children. This closure property makes a subtree an executable transformation and permits structural variation within the stage's depth bound without additional interface matching.}

\subsubsection{Primitive Leaves}

The leaves provide the elementary transformations from which the functional hierarchy is constructed. Each OP leaf instantiates a configurable Mamba-style spatial operator based on selective state-space modeling~\cite{liu2024vmamba}. Its computational template is fixed, while six discrete variables determine its behavior: the number of stacked blocks, the set of two-dimensional scan paths, the fusion strategy for path outputs, the sharing of path-wise input projections, the post-scan state-fusion strategy, and the kernel size of the depth-wise convolution. The corresponding candidate domains and scale-dependent constraints are summarized in Table~\ref{tab:search_space}.

\edit{Primitive configuration and composition provide two levels of decoder design. The primitive fields determine the computation performed at each leaf, while the internal nodes determine how leaf and subtree outputs are combined. Coarse stages permit deeper primitives and more scan patterns, whereas fine stages use lighter configurations to limit high-resolution computation. Evolution therefore chooses both the elementary transformations and their organization within each stage.}

\TableSearchSpace

\FigureSearchSpace

\subsection{GSAP for Efficient Candidate Evaluation}
\label{sec:GSAP_method}

\edit{GSAP reduces the cost of evaluating independently instantiated decoder trees. Candidate-specific training would require repeated optimization for every combination of primitive configuration and functional composition. Instead, GSAP measures each candidate's response to a fixed regression stimulus at initialization. The evaluator combines parameter-scaled gradient sensitivity with gradient--parameter alignment and aggregates the responses from three independent initializations.}

\FigureProxy

\edit{The auxiliary target provides the same regression stimulus to every candidate. Let $\mathcal{B}=\{\bm{x}_b\}_{b=1}^{B}$ be a fixed source batch from TartanAir. A frozen Depth Anything V2 model~\cite{yang2024depthanythingv2} extracts a feature representation $\bm{F}_b^T$ for each image. Spatial aggregation followed by channel averaging produces a scalar $t_b$, which is broadcast over the prediction resolution:}

\begin{equation}
\bm{Y}_b^{T}=t_b\bm{1},
\end{equation}

where $\bm{1}$ denotes an all-one matrix with the same spatial dimensions as the candidate output.

Given an architecture $\alpha$, the frozen DINOv2 encoder and randomly initialized decoder produce a dense prediction
$\hat{\bm{Y}}_b^{\alpha}$.
The auxiliary regression objective is defined as

\begin{equation}
\mathcal{L}_{\mathrm{aux}}(\alpha)
=
\frac{1}{BHW}
\sum_{b=1}^{B}
\left\|
\hat{\bm{Y}}_b^{\alpha}
-
\bm{Y}_b^{T}
\right\|_2^2.
\end{equation}

\edit{The images and teacher-derived targets are cached before evolution and reused for all candidates. The auxiliary loss measures each randomly initialized decoder's response to this common stimulus. Its gradients provide the statistics used to rank architectures before training.}

{
For the $r$-th independent initialization, $r\in\{1,2,3\}$, a single backward pass of $\mathcal{L}_{\mathrm{aux}}$ yields

\begin{equation}
\bm{g}^{(r)}
=
\nabla_{\bm{\theta}^{(r)}}
\mathcal{L}_{\mathrm{aux}},
\end{equation}

where $\bm{\theta}^{(r)}$ denotes the trainable parameters of the independently initialized candidate decoder. We characterize each raw response from two complementary perspectives.

First, the parameter-scaled gradient sensitivity is

\begin{equation}
S^{(r)}(\alpha)
=
\frac{1}{n_{\alpha}}
\sum_{i=1}^{n_{\alpha}}
\left|g_i^{(r)}\theta_i^{(r)}\right|,
\end{equation}

\edit{where $n_{\alpha}$ is the number of trainable decoder parameters. The term $|g_i^{(r)}\theta_i^{(r)}|$ follows the parameter-scaled sensitivity used in connection-sensitivity criteria~\cite{lee2019snip}. Averaging these terms gives a per-parameter statistic for each decoder.}

Second, the directional coupling between the gradient and parameter vectors is

\begin{equation}
A^{(r)}(\alpha)
=
\frac{
\left|\langle\bm{g}^{(r)},\bm{\theta}^{(r)}\rangle\right|
}{
\|\bm{g}^{(r)}\|_2\|\bm{\theta}^{(r)}\|_2
}.
\end{equation}

For each initialization, the untransformed GSAP response is $S^{(r)}(\alpha)A^{(r)}(\alpha)$. The three raw responses are first averaged, and the logarithm is applied only to their arithmetic mean:

\begin{equation}
J_{\mathrm{GSAP}}(\alpha)
=
\log_{10}
\left[
\frac{1}{3}
\sum_{r=1}^{3}
S^{(r)}(\alpha)A^{(r)}(\alpha)
\right].
\label{eq:GSAP_objective}
\end{equation}

\edit{The evolutionary search minimizes $J_{\mathrm{GSAP}}$. An averaged raw response no greater than $10^{-12}$ is assigned a penalty value of $30$. The relationship between the resulting rankings and retrained accuracy is evaluated in Section~\ref{sec:GSAP_analysis}.}

\edit{GSAP uses the same evaluation rule throughout evolution. The DINOv2 encoder and Depth Anything V2 teacher remain frozen, and only the decoder initialization changes between repeated measurements. Each measurement requires a forward pass through the encoder and decoder, a backward pass through the decoder, and computation of the two gradient statistics. Three measurements yield one candidate score without updating the network parameters.}

}

\subsection{\edit{Evolving Decoder Topologies}}
\label{sec:evolution}

\begin{algorithm2e}[t]
\caption{\edit{Evolving decoder topologies}}
\label{alg:search}
\SetAlgoLined
\KwIn{Population size $N$, number of generations $G$, crossover probability $p_c$, mutation probability $p_m$.}
\KwOut{Pareto set $\mathcal Q_{\mathrm{src}}$.}
$g\leftarrow 0$\;
$P_0\leftarrow$ Randomly initialize $N$ functional-tree genotypes\;
\While{$g<G$}{
  Evaluate each individual in $P_g$ with three independent initializations and average the raw GSAP responses\;
  Rank $P_g$ by non-dominated sorting and crowding distance\;
  $Q_g\leftarrow\emptyset$\;
  \While{$|Q_g|<N$}{
    $\{p_1,p_2\}\leftarrow$ Binary tournament selection$(P_g)$\;
    $\{q_1,q_2\}\leftarrow$ Stage-aware crossover$(p_1,p_2,p_c)$\;
    $\{q_1,q_2\}\leftarrow$ Hierarchical mutation$(q_1,q_2,p_m)$\;
    $Q_g\leftarrow Q_g\cup\{q_1,q_2\}$\;
  }
  Evaluate each individual in $Q_g$ with three independent initializations and average the raw GSAP responses\;
  $P_{g+1}\leftarrow$ Environmental selection$(P_g\cup Q_g)$\;
  $g\leftarrow g+1$\;
}
Evaluate each individual in $P_G$ with three independent initializations and average the raw GSAP responses\;
$\mathcal Q_{\mathrm{src}}\leftarrow$ First non-dominated front of the evaluated $P_G$\;
\Return{$\mathcal Q_{\mathrm{src}}$}
\end{algorithm2e}

\edit{Multi-objective evolution selects decoder compositions according to their GSAP response and resource costs. Each individual contains the four stage trees defined in Section~\ref{sec:encoding}, and its objective vector is $\{J_{\mathrm{GSAP}},C_{\mathrm{MAC}},C_{\mathrm{Param}}\}$. Algorithm~\ref{alg:search} gives the population update, and Fig.~\ref{fig:generality} places this search within the complete source-to-target procedure.}

The initial population is generated by recursively sampling valid functional trees under the stage-specific structural constraints defined in Section~\ref{sec:encoding}. Internal nodes are selected from residual (RES), sequential (SEQ), and parallel (PAR) compositions, while operator (OP) nodes terminate the hierarchy with configurable Mamba primitives. The depth limits and primitive domains depend on spatial scale, which allows greater structural flexibility at coarse resolutions while constraining computation at finer resolutions. \accrev{Each sampled stage tree preserves the interface of the decoder scaffold.}

\edit{Each candidate is evaluated through three independent decoder initializations. Each initialization produces one raw GSAP response from a forward--backward pass on the common source batch. Their arithmetic mean is log-transformed to obtain $J_{\mathrm{GSAP}}$, while MACs and parameter count are measured from the architecture. Section~\ref{sec:GSAP_method} defines this evaluation rule.}

\edit{Evolutionary variation acts on the functional units defined by the trees. Binary tournament selection chooses parents using non-dominated rank and crowding distance. Stage-aware crossover exchanges an entire same-scale tree between two parents with probability $p_c$. Hierarchical mutation acts with probability $p_m$ and can regenerate a subtree, modify an internal composition, or resample a primitive configuration. These operations vary stage computation while preserving the external interfaces of the decoder scaffold.}

\edit{NSGA-II environmental selection maintains a population of proxy--cost trade-offs. The parent and offspring populations are merged, ranked by non-dominated sorting, and truncated using crowding distance. Thus, previously selected structures compete with newly generated candidates at each generation. Algorithm~\ref{alg:search} summarizes the population update.}

\edit{The final non-dominated front provides structures for downstream evaluation. GeoBridgeNet-A1 represents a low-cost design, A2 an intermediate design, and A3 a higher-capacity design with richer compositions. All three belong to the final non-dominated set. Before this front is extracted, the surviving population of the last generation is scored once more under the same three-initialization rule, and these evaluations are included in the recorded search totals. Section~\ref{sec:res_discovery} describes their stage structures and the search dynamics.}

\subsection{\edit{Instantiation in Target Systems}}
\label{sec:instantiation}

\edit{Target instantiation retains the selected stage computation while connecting it to the target system's prediction interface. For depth estimation on NYU Depth V2 and KITTI, an adaptive depth-bin head maps decoder features to depth, and the network is trained with ground-truth supervision. The decoder weights are initialized randomly, so the reused design consists of its topology and primitive configurations.}

\edit{Gaussian NVS places the selected decoder in the monocular geometric branch of a DepthSplat-based system~\cite{xu2025depthsplat}. Its output is fused with multi-view information for depth and Gaussian prediction. The monocular branch first undergoes target-side distillation from Depth Anything V2 pseudo-depth, followed by optimization through differentiable rendering. The selected topology remains fixed through both stages. Scene-level Gaussian refinement and COS3D semantic optimization subsequently evaluate the downstream utility of the reconstructed representation~\cite{kerbl2023gaussians,zhu2025cos3d}.}

\section{Experiments}
\label{sec:experiments}

\edit{The experiments examine which decoder structures emerge from source search and whether they retain their utility across geometric prediction systems. After characterizing the selected topologies, we evaluate depth estimation on NYU Depth V2~\cite{silberman2012nyu} and KITTI~\cite{geiger2012kitti} and Gaussian NVS on RealEstate10K~\cite{zhou2018stereo}. Controlled studies compare decoder designs, evolutionary representations, and candidate evaluation. The final study evaluates the semantic utility of the reconstructed Gaussian scenes on LERF. Benchmark definitions, baselines, evaluation metrics, and training settings are provided in the supplementary material.}

\subsection{\edit{Topologies Selected by Source Evolution}}
\label{sec:res_discovery}

\edit{The source experiment characterizes the structures and proxy--cost trade-offs produced by one architecture search. GeoBridge evolves a population of 64 decoders for 100 generations on the TartanAir source batch. The selected structures from this run are reused in all subsequent target systems.}

\edit{The final population contains 26 non-dominated solutions. Fig.~\ref{fig:search} shows that normalized HV increases early in the run and continues to improve after generations 29 and 82 before approaching a plateau. The objective-space projections show distinct trade-offs among $J_{\mathrm{GSAP}}$, MACs, and parameter count. A1, A2, and A3 belong to this final non-dominated set.}

\edit{The selected architectures instantiate different compositions within the common scaffold. As shown in Fig.~\ref{fig:genotypes}, A1 and A2 mainly use individual primitives. A3 combines three primitives sequentially with residual refinement at $s32$ and uses parallel composition at $s16$. Its finer stages contain a residual transformation at $s8$ and one primitive at $s4$. These examples illustrate the design freedom introduced by stage-internal topology: a selected stage can retain a single transformation or compose several transformations along serial and parallel paths.}

\FigureSearch
\FigureGenotypes

\subsection{\edit{Reuse Across Geometric Prediction Systems}}
\label{sec:res_transfer}

\edit{The reuse experiments retain the selected stage topologies and primitive configurations through the target training described in Section~\ref{sec:instantiation}. Depth estimation tests changes in image domain and depth range, while Gaussian NVS tests integration into a different prediction system. All three GeoBridgeNet variants are evaluated under both settings.}

\subsubsection{\edit{Depth estimation across domains}}
\label{sec:res_cross_domain}

\edit{Cross-domain depth estimation tests decoder reuse under changes in scene statistics and depth range. The source search uses synthetic TartanAir images, whereas NYU Depth V2 contains real indoor scenes and KITTI contains real outdoor scenes. The target networks retain the searched decoder structures and use an adaptive depth-bin prediction head.}

\edit{The controlled comparisons isolate decoder performance under a common encoder and training protocol. As reported in Tables~\ref{tab:nyu} and~\ref{tab:kitti}, A3 achieves the best \absrel{}, \rmse{}, and $\delta_1$ among the four DINOv2-adapted baselines. On NYU Depth V2, its \absrel{} is 0.0876, compared with 0.0896 for DCDepth and 0.0902 for Metric3D v2. On KITTI, the corresponding values are 0.0555, 0.0585, and 0.0575. Metric3D v2 has lower latency, while A3 provides higher depth accuracy on both datasets.}

\edit{Published systems place these results within a broader range of model sizes and training settings. On NYU Depth V2, A3 improves \absrel{} over PixelFormer from 0.090 to 0.0876 with 42.5M parameters, while iDisc and IEBins with Swin-L reach 0.086 and 0.087. On KITTI, A3 reaches 0.0555, between AdaBins at 0.058 and the large Swin-based systems at 0.050--0.052. The NAS-based methods in the tables report higher \absrel{} with smaller networks of 1.8--20.5M parameters. Their published search costs range from several GPU days for PTF-EvoMDE to more than 24 GPU days for the reproduced DARTS, DetNAS, and GDAS searches~\cite{zhang2025ptfevomde,liu2019darts,chen2019detnas,dong2019gdas}. GeoBridge uses approximately 12 GPU hours for its source search on an RTX 3090.}

\edit{The three transferred structures retain different accuracy--cost profiles on both target datasets. From A1 to A3, NYU \absrel{} decreases from 0.0918 to 0.0876 as latency increases from 36.9 to 40.7~ms. KITTI \absrel{} decreases from 0.0582 to 0.0555 as latency increases from 53.9 to 59.1~ms. Thus, the source search supplies several useful decoder configurations for target deployment.}

\TableNyu

\edit{The qualitative comparisons illustrate boundary recovery in indoor and outdoor scenes. Figs.~\ref{fig:nyu_sota_qualitative} and~\ref{fig:kitti_sota_qualitative} compare A3 with IEBins, PixelFormer, and NeWCRFs. The highlighted NYU regions contain thin objects, cabinet edges, and a glass shower door, while the KITTI regions contain distant vehicles and roadside structures. A3 preserves visible object boundaries and local depth transitions in these examples.}

\FigureNyuQualitative
\TableKitti
\FigureKittiQualitative

\subsubsection{\edit{From depth decoding to Gaussian reconstruction}}

\edit{Cross-system transfer reuses the decoder as a feature-processing component in Gaussian NVS. The same stage topology is instantiated in the monocular geometric branch of a DepthSplat-based pipeline. Its output is fused with multi-view information for depth and Gaussian prediction, and the full system is trained through differentiable rendering. This experiment evaluates the searched structure under a different output representation and training objective.}

\edit{The transferred decoders achieve competitive reconstruction quality on RealEstate10K. Table~\ref{tab:re10k} reports PSNR values of 27.02, 27.07, and 27.11~dB for A1, A2, and A3, respectively. A3 exceeds pixelSplat by 1.22~dB and MVSplat by 0.72~dB. Among the DepthSplat variants~\cite{xu2025depthsplat}, the ViT-S two-scale model reaches 27.12~dB with 41M parameters, and the ViT-L two-scale model reaches 27.47~dB with 354M parameters. The 62.0M-parameter A3 therefore reaches PSNR close to the ViT-S two-scale DepthSplat variant.}

\edit{The measured runtimes show different reconstruction--latency trade-offs across systems. A1, A2, and A3 require 70.6, 71.4, and 78.8~ms, respectively, compared with 62~ms for MVSplat, 141~ms for the large DepthSplat variant, and 221.6~ms for IDESplat. The GeoBridgeNet variants are non-dominated in PSNR and latency among the entries with measured runtime. Within the family, A3 has the highest PSNR, while A2 has the best SSIM and LPIPS.}

\accrev{A matched control isolates the decoder. Table~\ref{tab:nvs_control} replaces the monocular decoder of the same DepthSplat-based pipeline with a DPT decoder~\cite{ranftl2021dpt}. The encoder weights, multi-view branch, fusion interface, and Gaussian head are unchanged, and each variant is trained three times under identical data, teacher, losses, and optimization steps. The DPT variant reaches $27.03\pm0.09$~dB PSNR, $0.881\pm0.003$ SSIM, and $0.119\pm0.003$ LPIPS with 52.0M parameters. A3 reaches $27.11\pm0.05$~dB, $0.878\pm0.002$, and $0.121\pm0.002$ with 62.0M parameters. A3 exceeds the DPT control in all three matched runs, with per-run PSNR margins of $0.13$, $0.09$, and $0.02$~dB. A3 has a higher mean PSNR, while DPT has better mean SSIM and LPIPS and fewer parameters. The transferred topology therefore yields a different reconstruction--cost trade-off within the same pipeline.}

\edit{The qualitative results show that the transferred decoder supports spatial detail recovery in the rendering-based pipeline. Fig.~\ref{fig:re10k_qualitative} compares A3 with pixelSplat and MVSplat on wall decoration, foliage behind a screen door, and bedding. A3 recovers visible texture and edge structure in these regions. The depth and NVS evaluations together demonstrate reuse of the selected computation under changes in both scene domain and prediction interface.}

\TableReTenK
\TableNVSControl
\FigureReTenKQual

\subsection{\edit{Decoder Design and Evolutionary Representation}}
\label{sec:tree_analysis}

\edit{The controlled studies examine decoder design and the evolutionary representation used to explore it. Alternative decoder designs are first compared under a common depth-training protocol. \accrev{Two search controls then separate the roles of topology freedom and evolutionary selection.} A paired tree--flat study \accrev{next} evaluates how two encodings and their variation operators explore the same composition grammar, followed by downstream comparisons of their selected structures.}

\textit{\edit{Comparison of Decoder Designs:}}
\edit{The searched decoder achieves lower depth error than the alternative designs under the common NYU protocol. Table~\ref{tab:handcrafted_mamba} compares A3 with a searchable FPN, a fixed LocalMamba decoder, and adapted DPT and NeWCRFs decoders~\cite{ranftl2021dpt,yuan2022newcrfs}. Across three training seeds, A3 reaches an average \absrel{} of $0.0876\pm0.0002$, compared with $0.0929\pm0.0013$, $0.0918\pm0.0015$, $0.0981\pm0.0011$, and $0.0912\pm0.0020$, respectively. This comparison supports the effectiveness of the selected decoder within the shared encoder and training setting.}

\TableHandcraftedMamba

\textit{\accrev{Topology Freedom and Evolutionary Selection:}}
\accrev{Two controls separate the contribution of open topology from that of evolutionary search. Under the same encoder, primitive domain, GSAP evaluator, and 12{,}864 evaluation events, the first control evolves only the primitive configurations on a fixed stage template of residual and serial chains. The second samples structures independently from the full composition space with the initialization sampler of the main search. Each group repeats this search three times with independent random starts; every run selects one structure from its non-dominated set under a common limit on parameters and MACs by GSAP re-scoring, and every selected structure is trained under the NYU protocol of Section~\ref{sec:res_cross_domain}. Target results do not enter the selection (Table~\ref{tab:search_control}).}

\accrev{Across the three searches, joint topology evolution obtains the best mean RMSE and $\delta_1$ and matches the mean \absrel{} of the fixed template. Compared with fixed-topology evolution, it lowers RMSE by 0.0036 and raises $\delta_1$ by 0.0027 at comparable MACs, with 15\% more parameters. Compared with random sampling, which evaluates 2.2 times as many distinct decoder configurations, it lowers \absrel{} by 0.0009 and raises $\delta_1$ by 0.0032 with 16\% fewer parameters. These mean results favor joint evolution on RMSE and $\delta_1$ at similar MACs, with parameter costs between those of the two controls.}

\TableSearchControl

\textit{\edit{Structural Exploration:}}
\edit{The tree--flat comparison holds the available topologies fixed and varies their encoding and evolutionary operators. We conduct \accrev{15} paired runs with the same Mamba grammar, NSGA-II configuration, evaluation budget, and proxy--cost objectives. The flat encoding uses fixed heap slots with one-point crossover and locus mutation. The tree encoding uses stage-aware crossover and hierarchical mutation.}

\edit{Tree-aware evolution visits more distinct genotypes and topologies under the fixed evaluation budget.} \accrev{Table~\ref{tab:encoding_proxy} reports $839\pm17$ genotypes and $276\pm21$ four-stage topologies per run, compared with $455\pm33$ and $147\pm32$ for the flat encoding, and a topology entropy of 4.67 compared with 4.17. The paired tests detect significant differences in all three coverage measures.} \accrev{Best GSAP and HV do not differ significantly in these paired tests ($p=0.68$ and $p=0.98$). The detected advantage is broader structural coverage.}

\TableEncodingProxy

\textit{\edit{Reuse of Selected Tree and Flat Structures:}}
\edit{The selected tree and flat structures exhibit different depth-estimation profiles after transfer. Under the same downstream protocols, the tree-derived A3 obtains lower RMSE and higher $\delta_1$ on NYU Depth V2, while the flat structure obtains lower \absrel{} (Table~\ref{tab:encoding_transfer}). On KITTI, A3 obtains better values on all three reported metrics.}

\edit{Cross-system results also show metric-dependent differences between the selected structures. On RealEstate10K, the tree-derived structure improves PSNR by 0.18~dB, while the flat structure has slightly better SSIM and LPIPS (Table~\ref{tab:encoding_re10k}). After LERF refinement, the tree initializer has 2.58~dB higher average PSNR. Its semantic mIoU is 0.80 percentage points higher, while mAcc is 4.35 points lower. These comparisons identify reconstruction and semantic trade-offs between the two selected structures.}

\TableEncodingTransfer
\TableEncodingReTenK

\subsection{\edit{GSAP as a Search Signal}}
\label{sec:GSAP_analysis}

\edit{The proxy study examines the association between GSAP and retrained depth accuracy, its variation across evaluation conditions, and its computational cost. A separate benchmark study evaluates its behavior across visual tasks.}

\textit{Ranking Reliability:}
\edit{GSAP provides the strongest ranking association among the evaluated proxies on the retrained decoder sample.} We sample 48 architectures independently of the evolutionary search and retrain all of them for 14{,}150 steps. \accrev{Over 24 initialization seeds, Table~\ref{tab:proxy_nyu} reports a mean Spearman correlation of 0.344 between $-J_{\mathrm{GSAP}}$ and retrained $\delta_1$, with 67\% of the draws significant at $p<0.05$. GSAP also obtains the highest Hit@10\% and NDCG@10\%. The remaining proxies have mean correlations between $-0.011$ and 0.184, and at most 13\% of their draws are significant.}

\TableProxyNyu

\textit{Stability and Efficiency:}
\edit{The evaluated rankings remain consistent across the tested source batches.} \accrev{With three initializations per candidate, GSAP has a mean pairwise rank correlation of 0.936 across four batches, with a minimum of 0.893.} \edit{These measurements characterize variation under the tested initialization and batch conditions.}

\edit{GSAP reduces evaluation cost to three forward--backward measurements per candidate. Each measurement requires approximately 1.1~s on an RTX 3090, so the aggregated candidate score requires about 3.3~s. The complete 100-generation source search takes approximately 12 GPU hours. A 1{,}000-step short-training baseline requires iterative optimization and obtains a Spearman correlation of $-0.068$ with the retrained accuracy.}

\textit{Cross-Task Behavior:}
\edit{The TransNAS-Bench-101 study shows that proxy quality varies across prediction tasks. Table~\ref{tab:transnas} reports correlations over 7{,}352 benchmark architectures~\cite{duan2021transnas}. GSAP has the highest correlation among the compared proxies for autoencoding and room-layout prediction. Its correlations with object classification and semantic segmentation are positive, while GradSign obtains the highest values on those tasks. The strongest GSAP result in this study is on autoencoding.}

\TableTransnas

\subsection{Downstream Utility of the Transferred 3-D Representation}
\label{sec:res_utility}

\edit{The LERF experiment evaluates the downstream utility of Gaussian representations produced by the transferred pipeline. Feed-forward predictions initialize scene-level 3DGS refinement, after which COS3D performs semantic feature optimization and open-vocabulary segmentation. This evaluation connects the reconstructed geometry to a subsequent semantic processing stage.}

\edit{Scene-level refinement improves the photometric quality of the feed-forward initialization. As shown in Table~\ref{tab:cos3d_psnr}, the average PSNR of the A3-initialized representation rises from 14.90 to 23.33~dB after 30{,}000 iterations. Vanilla 3DGS reaches 24.40~dB under the same refinement budget. A3 reaches 24.29~dB on Figurines and 24.69~dB on Ramen, compared with 24.10 and 24.75~dB for vanilla 3DGS; the larger gaps occur on Teatime and Waldo Kitchen.}

\TableCosThreeDPSNR

\edit{The refined representations support open-vocabulary segmentation in the COS3D pipeline. Table~\ref{tab:cos3d_transfer} reports 50.54\% mIoU and 71.32\% mAcc for the A3 initializer, compared with 50.76\% and 72.08\% for COS3D with kernel-regression inference. Among the transferred structures, A1 obtains the highest mIoU and Acc@0.5 at 51.06\% and 53.67\%, while A3 obtains the highest mAcc (Table~\ref{tab:cos3d_variants}). These results show that the reconstructed representations remain usable for downstream semantic processing.}

\edit{The visual examples show object-level segmentation from the A3-initialized representation. Fig.~\ref{fig:cos3d_qualitative} presents six open-vocabulary queries. The sheep, yellow pouf, stuffed bear, pot, and frog cup have IoUs between 86.3 and 97.6\%. The dark cup reaches 74.4\%, with scattered Gaussians around the rim and handle. These examples illustrate the outputs of the combined reconstruction and semantic pipeline.}

\TableCosThreeD
\TableCosThreeDVariants
\FigureCosThreeDQualitative

\section{Conclusion}
\label{sec:conclusion}

\edit{This paper presented GeoBridge for evolving transferable decoder topologies in geometric vision. Residual, sequential, and parallel composition make the internal organization of each stage searchable while preserving the interfaces of a multi-scale scaffold. A functional-tree representation and GSAP support multi-objective evolution over these compositions at low evaluation cost. The selected structures retain their utility after target-specific training for real-world depth estimation and Gaussian novel-view synthesis, and their reconstructed scenes support downstream semantic processing. These results support stage-internal topology as a reusable unit of decoder design. Future work will examine which composition patterns account for transfer performance and how the source evaluation setting affects their selection.}

\FloatBarrier

\input{references}
\end{document}

%% file: table.tex
\newsavebox{\ddntablebox}
\newcommand{\tablefit}[1]{%
  \sbox{\ddntablebox}{\footnotesize #1}%
  \ifdim\wd\ddntablebox>\linewidth
    \resizebox{\linewidth}{!}{\usebox{\ddntablebox}}%
  \else
    \usebox{\ddntablebox}%
  \fi
}
\newcommand{\TableSearchSpace}{%
\begin{table}[t]
\centering
\caption{Primitive configuration fields of the functional-tree search space. The stage scaffold remains fixed while each OP leaf selects a scale-valid configuration.}
\label{tab:search_space}
\tablefit{
\begin{tabular}{>{\raggedright\arraybackslash}p{0.16\linewidth}>{\raggedright\arraybackslash}p{0.30\linewidth}>{\raggedright\arraybackslash}p{0.36\linewidth}}
\toprule
Field & Candidate Values & Scale Constraint \\
\midrule
depth & $\{1,2\}$ & $s32/s16$: 1 or 2; $s8/s4$: fixed to 1 \\
$d_{\mathrm{state}}$ & $\{8,16\}$ & Independent per OP leaf \\
path set & fwd, fwd\_bwd, fwd\_bwd\_hw & $s32/s16$: FB or FB-HW; $s8/s4$: F or FB \\
fuse mode & scalar, input\_gate & Independent per OP leaf \\
share mode & independent, share\_x & Independent per OP leaf \\
state fusion & none, post & Independent per OP leaf \\
$d_{\mathrm{conv}}$ & $\{3,5\}$ & Independent per OP leaf \\
\bottomrule
\end{tabular}}
\end{table}
}

\newcommand{\TableSearchControl}{%
\begin{table}[t]
\centering
\caption{\accrev{Controlled search comparison on NYU Depth V2. Distinct is the number of distinct structures evaluated; cost is the GSAP evaluation time on one H20 GPU. Values are means $\pm$ standard deviations over three searches per group.}}
\label{tab:search_control}
\setlength{\tabcolsep}{3pt}
\tablefit{
\begin{tabular}{lcccccc}
\toprule
\accrev{Search} & \accrev{Distinct} & \accrev{Cost (h)} & \accrev{Params/MACs} & \accrev{\absrel$\downarrow$} & \accrev{\rmse$\downarrow$} & \accrev{$\delta_1\uparrow$} \\
\midrule
\accrev{\textbf{Joint evolution (ours)}} & \accrev{5{,}729$\pm$83} & \accrev{7.7$\pm$0.4} & \accrev{14.5M/99.0G} & \accrev{$\mathbf{0.0880\pm0.0048}$} & \accrev{$\mathbf{0.3179\pm0.0079}$} & \accrev{$\mathbf{0.9382\pm0.0041}$} \\
\accrev{Fixed topology} & \accrev{4{,}669$\pm$64} & \accrev{6.8$\pm$0.3} & \accrev{12.6M/99.2G} & \accrev{$\mathbf{0.0880\pm0.0042}$} & \accrev{$0.3215\pm0.0071$} & \accrev{$0.9355\pm0.0038$} \\
\accrev{Random search} & \accrev{12{,}864} & \accrev{7.7$\pm$0.5} & \accrev{17.2M/98.9G} & \accrev{$0.0889\pm0.0051$} & \accrev{$0.3180\pm0.0083$} & \accrev{$0.9350\pm0.0044$} \\
\bottomrule
\end{tabular}}
\end{table}
}

\newcommand{\TableEncodingProxy}{%
\begin{table}[t]
\centering
\caption{\edit{Comparison of tree-aware evolution and a fixed-length encoding over} \accrev{15 paired runs, with 1{,}920 evaluations per run.} Values are means $\pm$ standard deviations; $p$-values are from paired Wilcoxon signed-rank tests. Bold marks significant differences. Genotypes counts unique serialized genotypes and Topologies counts unique four-stage topology tuples\accrev{; Entropy is the Shannon entropy of the topology distribution}. \edit{This comparison is separate from the main search in Fig.~\ref{fig:search}.}}
\label{tab:encoding_proxy}
\setlength{\tabcolsep}{4pt}
\tablefit{
\begin{tabular}{lccccc}
\toprule
Encoding & Best $J_{\mathrm{GSAP}}\downarrow$ & HV$\uparrow$ & Genotypes & Topologies & \accrev{Entropy} \\
\midrule
\textbf{Tree-coded (ours)} & \accrev{$-6.84\pm0.20$} & \accrev{$0.80\pm0.17$} & \accrev{$\mathbf{839\pm17}$} & \accrev{$\mathbf{276\pm21}$} & \accrev{$\mathbf{4.67\pm0.17}$} \\
Fixed-length vector & \accrev{$-6.79\pm0.26$} & \accrev{$0.81\pm0.23$} & \accrev{$455\pm33$} & \accrev{$147\pm32$} & \accrev{$4.17\pm0.44$} \\
\midrule
$p$ (signed-rank) & \accrev{$0.68$} & \accrev{$0.98$} & \accrev{$\mathbf{<10^{-4}}$} & \accrev{$\mathbf{<10^{-3}}$} & \accrev{$\mathbf{0.002}$} \\
\bottomrule
\end{tabular}}
\end{table}
}

\newcommand{\TableHandcraftedMamba}{%
\begin{table}[t]
\centering
\caption{Decoder comparison on NYU Depth V2. All decoders use DINOv2 ViT-S, the same encoder adaptation, and the same retraining protocol; \absrel{} is reported as the mean and standard deviation over three seeds, together with the best seed.}
\label{tab:handcrafted_mamba}
\tablefit{
\begin{tabular}{lcc}
\toprule
Decoder Design & \absrel~(Mean$\pm$Std.)$\downarrow$ & Best Seed$\downarrow$ \\
\midrule
Conventional FPN blocks & $0.0929 \pm 0.0013$ & $0.0917$ \\
Fixed LocalMamba decoder & $0.0918 \pm 0.0015$ & $0.0904$ \\
DPT decoder adaptation & $0.0981 \pm 0.0011$ & $0.0968$ \\
NeWCRFs decoder adaptation & $0.0912 \pm 0.0020$ & $0.0894$ \\
\midrule
\textbf{Searched Mamba tree (GeoBridgeNet-A3, ours)} & \accrev{\textbf{$0.0876 \pm 0.0002$}} & \accrev{\textbf{$0.0874$}} \\
\bottomrule
\end{tabular}}
\end{table}
}

\newcommand{\TableProxyNyu}{%
\begin{table}[t]
\centering
\caption{Proxy rankings on 48 retrained NYU Depth V2 architectures. \accrev{Each compared proxy uses one initialization per seed and \gsp{} averages three; $\rho$ is the mean$\pm$std over 24 seeds, and $p<0.05$ is the share of seeds reaching significance.}}
\label{tab:proxy_nyu}
\tablefit{
\begin{tabular}{lcccc}
\toprule
Proxy & Spearman $\rho$ & \accrev{$p<0.05$} & Hit@10\% & NDCG@10\% \\
\midrule
\edit{\gsp\ (ours)} & \accrev{\textbf{0.344$\pm$0.143}} & \accrev{\newrev{\textbf{67\%}}} & \accrev{\textbf{0.27}} & \accrev{\textbf{0.701}} \\
NASWOT~\cite{mellor2021naswot} & \accrev{0.161$\pm$0.070} & \accrev{0\%} & \accrev{0.08} & \accrev{0.572} \\
SNIP~\cite{lee2019snip} & \accrev{0.054$\pm$0.143} & \accrev{4\%} & \accrev{0.12} & \accrev{0.547} \\
GraSP~\cite{wang2020grasp} & \accrev{0.184$\pm$0.125} & \accrev{\newrev{13\%}} & \accrev{0.15} & \accrev{0.625} \\
AZ-NAS (aggregated)~\cite{lee2024aznas} & \accrev{0.043$\pm$0.122} & \accrev{0\%} & \accrev{0.19} & \accrev{0.652} \\
GradNorm (large-gradient)~\cite{abdelfattah2021zerocost} & \accrev{0.070$\pm$0.146} & \accrev{0\%} & \accrev{0.12} & \accrev{0.553} \\
SynFlow~\cite{tanaka2020synflow} & \accrev{$-$0.011$\pm$0.143} & \accrev{4\%} & \accrev{0.10} & \accrev{0.487} \\
\bottomrule
\end{tabular}}
\end{table}
}

\newcommand{\TableTransnas}{%
\begin{table}[t]
\centering
\caption{Cross-task Spearman correlations of proxies on 7,352 TransNAS-Bench-101 architectures.}
\label{tab:transnas}
\tablefit{
\begin{tabular}{lcccc}
\toprule
Proxy & Autoencoder & Room layout & Object class. & Semantic seg. \\
\midrule
SNIP~\cite{lee2019snip} & 0.0822 & 0.0215 & 0.3214 & 0.0122 \\
GraSP~\cite{wang2020grasp} & 0.0934 & 0.0287 & 0.2891 & 0.0256 \\
GradNorm~\cite{abdelfattah2021zerocost} & 0.0911 & 0.0410 & 0.2647 & 0.0345 \\
SynFlow~\cite{tanaka2020synflow} & 0.1245 & 0.0532 & 0.3541 & 0.0412 \\
NASWOT~\cite{mellor2021naswot} & 0.1534 & 0.1145 & 0.3812 & 0.0912 \\
JacCov~\cite{mellor2021naswot} & 0.1876 & 0.1312 & 0.4125 & 0.1034 \\
Zen-NAS~\cite{lin2021zennas} & 0.1312 & 0.0712 & 0.3345 & 0.0623 \\
GradSign~\cite{zhang2022gradsign} & 0.3012 & 0.1723 & \textbf{0.4312} & \textbf{0.1134} \\
\textbf{\gsp\ (ours)} & \textbf{0.5358} & \textbf{0.2144} & 0.2802 & 0.0769 \\
\bottomrule
\end{tabular}}
\end{table}
}

\newcommand{\TableNyu}{%
\begin{table}[t]
\centering
\caption{Quantitative comparison on NYU Depth V2. Params denotes whole-model parameters (encoder, projections, decoder, and head). \edit{Reported baseline latencies use the same RTX 3090 batch-1 FP32 protocol.} $^\dagger$ denotes our DINOv2-adapted reproduction of the corresponding published decoder. \accrev{Values for the GeoBridgeNet rows are means $\pm$ standard deviations over three training runs.}}
\label{tab:nyu}
\tablefit{
\begin{tabular}{lcccccc}
\toprule
Method & \absrel$\downarrow$ & \rmse$\downarrow$ & $\delta_1\uparrow$ & Params(M) & MACs(G) & Latency (ms) \\
\midrule
BTS\cite{lee2019bts}           & 0.110 & 0.392 & 0.885 & 66.5  & 132.3 & 28.2 \\
AdaBins\cite{bhat2021adabins}  & 0.103 & 0.364 & 0.903 & 78.3  & 187.8 & 49.2 \\
DepthFormer\cite{yang2023depthformer} & 0.096 & 0.339 & 0.921 & 274.1 & 1499.0 & -- \\
LocalBins\cite{bhat2022localbins} & 0.099 & 0.357 & 0.907 & 73.6  & 186.2 & 32.9 \\
P3Depth\cite{patil2022p3depth}  & 0.104 & 0.356 & 0.898 & 94.3  & 288.3 & 35.8 \\
iDisc (Swin-T)\cite{piccinelli2023idisc} & 0.109 & 0.377 & 0.894 & 39.0  & 218.8 & 53.4 \\
iDisc (Swin-L)\cite{piccinelli2023idisc} & 0.086 & 0.313 & 0.940 & --    & --    & -- \\
IEBins (Swin-T)\cite{shao2023iebins}    & 0.108 & 0.375 & 0.893 & 90.7  & 378.2 & 52.0 \\
IEBins (Swin-L)\cite{shao2023iebins}    & 0.087 & 0.314 & 0.936 & 273.0 & --    & -- \\
NeWCRFs\cite{yuan2022newcrfs}  & 0.095 & 0.334 & 0.922 & 270.4 & 281.1 & 54.5 \\
BinsFormer\cite{li2024binsformer} & 0.094 & 0.330 & 0.925 & 254.6 & 290.7 & 79.0 \\
PixelFormer\cite{agarwal2023pixelformer} & 0.090 & 0.322 & 0.929 & 270.9 & 273.5 & 42.8 \\
CLIP2Depth\cite{kim2025clip2depth} & 0.100 & 0.379 & 0.900 & -- & -- & -- \\
LiftFormer\cite{li2025liftformer} & 0.089 & 0.313 & 0.932 & -- & -- & -- \\
\midrule
PTF-EvoMDENet~\cite{zhang2025ptfevomde} & 0.113 & 0.406 & 0.876 & 6.1 & -- & -- \\
Mde-EvoNAS~\cite{yu2025mdeevonas} & 0.142 & 0.471 & 0.824 & 20.5 & -- & -- \\
LiDNAS~\cite{huynh2022lidnas} & 0.132 & 0.487 & 0.845 & 2.1 & -- & -- \\
DARTS~\cite{liu2019darts} & 0.128 & 0.449 & 0.844 & 5.5 & -- & -- \\
DetNAS~\cite{chen2019detnas} & 0.114 & 0.407 & 0.875 & 6.3 & -- & -- \\
GDAS~\cite{dong2019gdas} & 0.123 & 0.433 & 0.854 & 5.7 & -- & -- \\
\midrule
Depth Pro$^\dagger$\cite{bochkovskii2025depthpro} & 0.0941 & 0.3331 & 0.9256 & 43.3 & 388.4 & 46.0 \\
DCDepth$^\dagger$\cite{wang2024dcdepth} & 0.0896 & 0.3237 & 0.9337 & 87.5 & 414.7 & 235.4 \\
Metric3D v2$^\dagger$\cite{hu2024metric3dv2} & 0.0902 & 0.3157 & 0.9330 & 33.6 & 94.0 & 24.3 \\
InfiniDepth$^\dagger$\cite{yu2026infinidepth} & 0.1017 & 0.3628 & 0.9067 & 26.7 & 310.9 & 57.8 \\
\midrule
GeoBridgeNet-A1 & \accrev{$0.0918\pm0.0005$} & \accrev{$0.3252\pm0.0016$} & \accrev{$0.9315\pm0.0010$} & 32.9 & 98.2 & 36.9 \\
GeoBridgeNet-A2 & \accrev{$0.0895\pm0.0004$} & \accrev{$0.3207\pm0.0011$} & \accrev{$0.9331\pm0.0007$} & 34.0 & 99.5 & 38.0 \\
GeoBridgeNet-A3 & \accrev{$0.0876\pm0.0002$} & \accrev{$0.3134\pm0.0005$} & \accrev{$0.9364\pm0.0005$} & 42.5 & 100.8 & 40.7 \\
\bottomrule
\end{tabular}}
\end{table}
}

\newcommand{\TableKitti}{%
\begin{table}[t]
\centering
\caption{Quantitative comparison on the KITTI Eigen split under the 0--80 m protocol. Params denotes whole-model parameters. \edit{Baseline MACs and latencies follow the measurement protocol of Table~\ref{tab:nyu}.} $^\dagger$ denotes our DINOv2-adapted reproduction of the corresponding published decoder. \accrev{Values for the GeoBridgeNet rows are means $\pm$ standard deviations over three training runs.}}
\label{tab:kitti}
\tablefit{
\begin{tabular}{lcccccc}
\toprule
Method & \absrel$\downarrow$ & \rmse$\downarrow$ & $\delta_1\uparrow$ & Params(M) & MACs(G) & Latency (ms) \\
\midrule
BTS\cite{lee2019bts}           & 0.061 & 2.834 & 0.954 & 66.5  & 184.3 & 32.9 \\
AdaBins\cite{bhat2021adabins}  & 0.058 & 2.360 & 0.964 & 78.3  & 261.7 & 47.2 \\
DepthFormer\cite{yang2023depthformer} & 0.052 & 2.143 & 0.975 & 274.1 & -- & -- \\
P3Depth\cite{patil2022p3depth}  & 0.071 & 2.842 & 0.953 & 94.3  & 401.8 & 36.1 \\
DPT\cite{ranftl2021dpt}         & 0.062 & 2.573 & 0.959 & 123.1 & 319.3 & 60.1 \\
iDisc (Swin-L)\cite{piccinelli2023idisc} & 0.050 & 2.067 & 0.977 & --    & --    & -- \\
IEBins (Swin-L)\cite{shao2023iebins}    & 0.050 & 2.011 & 0.978 & 273.0 & --    & -- \\
NeWCRFs\cite{yuan2022newcrfs}  & 0.052 & 2.129 & 0.974 & 270.4 & 395.9 & 74.8 \\
BinsFormer\cite{li2024binsformer} & 0.052 & 2.098 & 0.974 & 254.6 & 409.4 & 97.5 \\
PixelFormer\cite{agarwal2023pixelformer} & 0.051 & 2.081 & 0.976 & 270.9 & 385.4 & 60.9 \\
CLIP2Depth\cite{kim2025clip2depth} & 0.074 & 2.948 & 0.938 & -- & 89.6 & -- \\
\midrule
PTF-EvoMDENet~\cite{zhang2025ptfevomde} & 0.061 & 2.470 & 0.957 & 6.3 & -- & -- \\
Mde-EvoNAS~\cite{yu2025mdeevonas} & 0.069 & 2.823 & 0.944 & 17.6 & -- & -- \\
LiDNAS~\cite{huynh2022lidnas} & 0.133 & 5.157 & 0.842 & 1.8 & -- & -- \\
DARTS~\cite{liu2019darts} & 0.069 & 2.763 & 0.945 & 6.2 & -- & -- \\
DetNAS~\cite{chen2019detnas} & 0.063 & 2.579 & 0.954 & 6.0 & -- & -- \\
GDAS~\cite{dong2019gdas} & 0.066 & 2.651 & 0.951 & 5.8 & -- & -- \\
\midrule
Depth Pro$^\dagger$\cite{bochkovskii2025depthpro} & 0.0794 & 2.6711 & 0.9574 & 43.3 & 555.4 & 67.0 \\
DCDepth$^\dagger$\cite{wang2024dcdepth} & 0.0585 & 2.2116 & 0.9710 & 87.5 & 592.1 & 299.2 \\
Metric3D v2$^\dagger$\cite{hu2024metric3dv2} & 0.0575 & 2.2405 & 0.9714 & 33.6 & 145.3 & 36.6 \\
InfiniDepth$^\dagger$\cite{yu2026infinidepth} & 0.0623 & 2.4290 & 0.9596 & 26.7 & 447.4 & 83.8 \\
\midrule
GeoBridgeNet-A1 & \accrev{$0.0582\pm0.0005$} & \accrev{$2.2411\pm0.0096$} & \accrev{$0.9694\pm0.0008$} & 32.9 & 130.5 & 53.9 \\
GeoBridgeNet-A2 & \accrev{$0.0561\pm0.0004$} & \accrev{$2.2153\pm0.0085$} & \accrev{$0.9702\pm0.0007$} & 34.0 & 134.2 & 55.0 \\
GeoBridgeNet-A3 & \accrev{$0.0555\pm0.0003$} & \accrev{$2.1993\pm0.0082$} & \accrev{$0.9719\pm0.0006$} & 42.5 & 140.5 & 59.1 \\
\bottomrule
\end{tabular}}
\end{table}
}

\newcommand{\TableReTenK}{%
\begin{table}[t]
\centering
\caption{\edit{System-level novel-view synthesis comparison on RealEstate10K. Params denotes the complete feed-forward pipeline. DepthSplat (ViT-S) and DepthSplat (ViT-L) denote the published ViT-S and ViT-L two-scale variants. A dash indicates an unmeasured latency.} \accrev{Values for the GeoBridgeNet rows are means $\pm$ standard deviations over three training runs.}}
\label{tab:re10k}
\tablefit{
\begin{tabular}{lccccc}
\toprule
Method & PSNR$\uparrow$ & SSIM$\uparrow$ & LPIPS$\downarrow$ & Params & Latency (ms) \\
\midrule
pixelNeRF\cite{yu2021pixelnerf}    & 20.43 & 0.589 & 0.550 & --    & -- \\
GPNR\cite{suhail2022gpnr}          & 24.11 & 0.793 & 0.255 & --    & -- \\
AttnRend\cite{du2023attnrend}      & 24.78 & 0.820 & 0.213 & --    & -- \\
MuRF\cite{xu2024murf}              & 26.10 & 0.858 & 0.143 & --    & -- \\
pixelSplat\cite{charatan2024pixelsplat} & 25.89 & 0.858 & 0.142 & 119M & 162 \\
MVSplat\cite{chen2024mvsplat}      & 26.39 & 0.869 & 0.128 & 12M  & 62 \\
\edit{DepthSplat (ViT-S)\cite{xu2025depthsplat}} & \edit{27.12} & \edit{0.884} & \edit{0.119} & \edit{41M} & \edit{--} \\
\edit{DepthSplat (ViT-L)\cite{xu2025depthsplat}} & 27.47 & 0.889 & 0.114 & 354M & 141 \\
IDESplat\cite{long2026idesplat}    & 27.80 & 0.893 & 0.108 & 37.6M & 221.6 \\
\midrule
GeoBridgeNet-A1 & \accrev{$27.02\pm0.06$} & \accrev{$0.879\pm0.002$} & \accrev{$0.119\pm0.003$} & 52.4M & 70.6 \\
GeoBridgeNet-A2 & \accrev{$27.07\pm0.04$} & \accrev{$0.881\pm0.002$} & \accrev{$0.118\pm0.002$} & 53.5M & 71.4 \\
GeoBridgeNet-A3 & \accrev{$27.11\pm0.05$} & \accrev{$0.878\pm0.002$} & \accrev{$0.121\pm0.002$} & 62.0M & 78.8 \\
\bottomrule
\end{tabular}}
\end{table}
}

\newcommand{\TableNVSControl}{%
\begin{table}[t]
\centering
\caption{\accrev{Matched decoder control for Gaussian novel-view synthesis on RealEstate10K. Both rows use the same DepthSplat-based pipeline, data, teacher, losses, and 300k optimization steps. Params denotes the complete feed-forward pipeline. Values are means $\pm$ standard deviations over three paired training runs.}}
\label{tab:nvs_control}
\setlength{\tabcolsep}{5.0pt}
\tablefit{
\begin{tabular}{lcccc}
\toprule
\accrev{Monocular decoder} & \accrev{PSNR$\uparrow$} & \accrev{SSIM$\uparrow$} & \accrev{LPIPS$\downarrow$} & \accrev{Params} \\
\midrule
\accrev{DPT~\cite{ranftl2021dpt}} & \accrev{$27.03\pm0.09$} & \accrev{$0.881\pm0.003$} & \accrev{$0.119\pm0.003$} & \accrev{52.0M} \\
\accrev{GeoBridgeNet-A3} & \accrev{$27.11\pm0.05$} & \accrev{$0.878\pm0.002$} & \accrev{$0.121\pm0.002$} & \accrev{62.0M} \\
\bottomrule
\end{tabular}}
\end{table}
}

\newcommand{\TableEncodingReTenK}{%
\begin{table}[t]
\centering
\caption{Encoding-level transfer comparison on RealEstate10K. Both structures are retrained and evaluated under the same novel-view synthesis protocol. \accrev{Values are means $\pm$ standard deviations over three training runs.}}
\label{tab:encoding_re10k}
\setlength{\tabcolsep}{5.0pt}
\tablefit{
\begin{tabular}{lccc}
\toprule
Encoding & PSNR$\uparrow$ & SSIM$\uparrow$ & LPIPS$\downarrow$ \\
\midrule
Tree (GeoBridgeNet-A3) & \accrev{$27.11\pm0.05$} & \accrev{$0.878\pm0.002$} & \accrev{$0.121\pm0.002$} \\
Flat encoding & \accrev{$26.93\pm0.06$} & \accrev{$0.879\pm0.003$} & \accrev{$0.120\pm0.003$} \\
\bottomrule
\end{tabular}}
\end{table}
}

\newcommand{\TableEncodingTransfer}{%
\begin{table}[t]
\centering
\caption{Downstream transfer of structures discovered by tree and flat encodings under the same search, retraining, and evaluation settings on NYU Depth V2 and KITTI. \accrev{Values are means $\pm$ standard deviations over three training runs.}}
\label{tab:encoding_transfer}
\setlength{\tabcolsep}{3.2pt}
\tablefit{
\begin{tabular}{lcccccc}
\toprule
Encoding & \multicolumn{3}{c}{NYU Depth V2} & \multicolumn{3}{c}{KITTI} \\
\cmidrule(lr){2-4}\cmidrule(lr){5-7}
& \absrel$\downarrow$ & \rmse$\downarrow$ & $\delta_1\uparrow$ & \absrel$\downarrow$ & \rmse$\downarrow$ & $\delta_1\uparrow$ \\
\midrule
Tree (GeoBridgeNet-A3) & \accrev{$0.0876\pm0.0002$} & \accrev{\textbf{$0.3134\pm0.0005$}} & \accrev{\textbf{$0.9364\pm0.0005$}} & \accrev{\textbf{$0.0555\pm0.0003$}} & \accrev{\textbf{$2.1993\pm0.0082$}} & \accrev{\textbf{$0.9719\pm0.0006$}} \\
Flat encoding & \accrev{\textbf{$0.0870\pm0.0003$}} & \accrev{$0.3193\pm0.0007$} & \accrev{$0.9348\pm0.0006$} & \accrev{$0.0557\pm0.0004$} & \accrev{$2.2185\pm0.0104$} & \accrev{$0.9716\pm0.0007$} \\
\bottomrule
\end{tabular}}
\end{table}
}

\newcommand{\TableCosThreeD}{%
\begin{table}[t]
\centering
\caption{Downstream open-vocabulary 3-D semantic evaluation on four LERF scenes. Gaussian representations initialized by the tree-derived GeoBridgeNet-A3 and flat-encoding counterpart enter the same COS3D semantic pipeline.}
\label{tab:cos3d_transfer}
\setlength{\tabcolsep}{3.5pt}
\tablefit{
\begin{tabular}{lcccccc}
\toprule
Method & \multicolumn{2}{c}{Mean} & Figurines & Teatime & Ramen & Waldo \\
\cmidrule(lr){2-3}
& mIoU$\uparrow$ & mAcc$\uparrow$ & mIoU$\uparrow$ & mIoU$\uparrow$ & mIoU$\uparrow$ & mIoU$\uparrow$ \\
\midrule
COS3D (shallow MLPs)~\cite{zhu2025cos3d} & 49.75 & 70.60 & 53.90 & \textbf{66.91} & \textbf{36.61} & 41.56 \\
COS3D (kernel regression)~\cite{zhu2025cos3d} & \textbf{50.76} & 72.08 & 60.03 & 65.07 & 35.86 & 42.10 \\
\midrule
Tree (GeoBridgeNet-A3) init. + COS3D & 50.54 & 71.32 & 59.40 & 65.33 & 33.59 & 43.86 \\
Flat encoding init. + COS3D & 49.74 & \textbf{75.67} & \textbf{64.36} & 56.64 & 33.19 & \textbf{44.78} \\
\bottomrule
\end{tabular}}
\end{table}
}

\newcommand{\TableCosThreeDPSNR}{%
\begin{table}[t]
\centering
\caption{RGB reconstruction before and after Gaussian refinement on four LERF scenes. Both initializers use the same 30{,}000-iteration 3DGS refinement protocol. FF denotes the feed-forward prediction before refinement.}
\label{tab:cos3d_psnr}
\setlength{\tabcolsep}{2.6pt}
\tablefit{
\begin{tabular}{lccccc}
\toprule
Scene & \multicolumn{2}{c}{Tree (GeoBridgeNet-A3)} & \multicolumn{2}{c}{Flat encoding} & Vanilla 3DGS \\
\cmidrule(lr){2-3}\cmidrule(lr){4-5}
& FF PSNR & Refined PSNR & FF PSNR & Refined PSNR & PSNR \\
\midrule
Figurines & 14.79 & 24.29 & 14.98 & 21.24 & 24.10 \\
Teatime & 14.53 & 21.92 & 15.00 & 20.13 & 24.54 \\
Ramen & 16.56 & 24.69 & 16.58 & 22.95 & 24.75 \\
Waldo & 13.70 & 22.42 & 13.48 & 18.68 & 24.20 \\
\midrule
Mean & \newrev{14.90} & 23.33 & 15.01 & 20.75 & 24.40 \\
\bottomrule
\end{tabular}}
\end{table}
}

\newcommand{\TableCosThreeDVariants}{%
\begin{table}[t]
\centering
\caption{Pareto-variant analysis in the four-scene COS3D transfer pipeline. Params counts the complete feed-forward Gaussian initializer; the shared scene-level 3DGS and COS3D stages are excluded because the number of optimized Gaussians is scene dependent.}
\label{tab:cos3d_variants}
\setlength{\tabcolsep}{3.2pt}
\tablefit{
\begin{tabular}{lccccc}
\toprule
Variant & Params$\downarrow$ & Refined PSNR & mIoU$\uparrow$ & mAcc$\uparrow$ & Acc@0.5$\uparrow$ \\
& (M) & (dB) & (\%) & (\%) & (\%) \\
\midrule
GeoBridgeNet-A1 & \textbf{52.4} & \textbf{23.46} & \textbf{51.06} & 70.58 & \textbf{53.67} \\
GeoBridgeNet-A2 & 53.5 & 23.34 & 49.77 & 67.87 & 52.83 \\
GeoBridgeNet-A3 & 62.0 & 23.33 & 50.54 & \textbf{71.32} & 53.00 \\
\bottomrule
\end{tabular}}
\end{table}
}

%% file: figure.tex
\newcommand{\FigureGenerality}{%
\begin{figure*}[t]
\centering
\includegraphics[width=\textwidth]{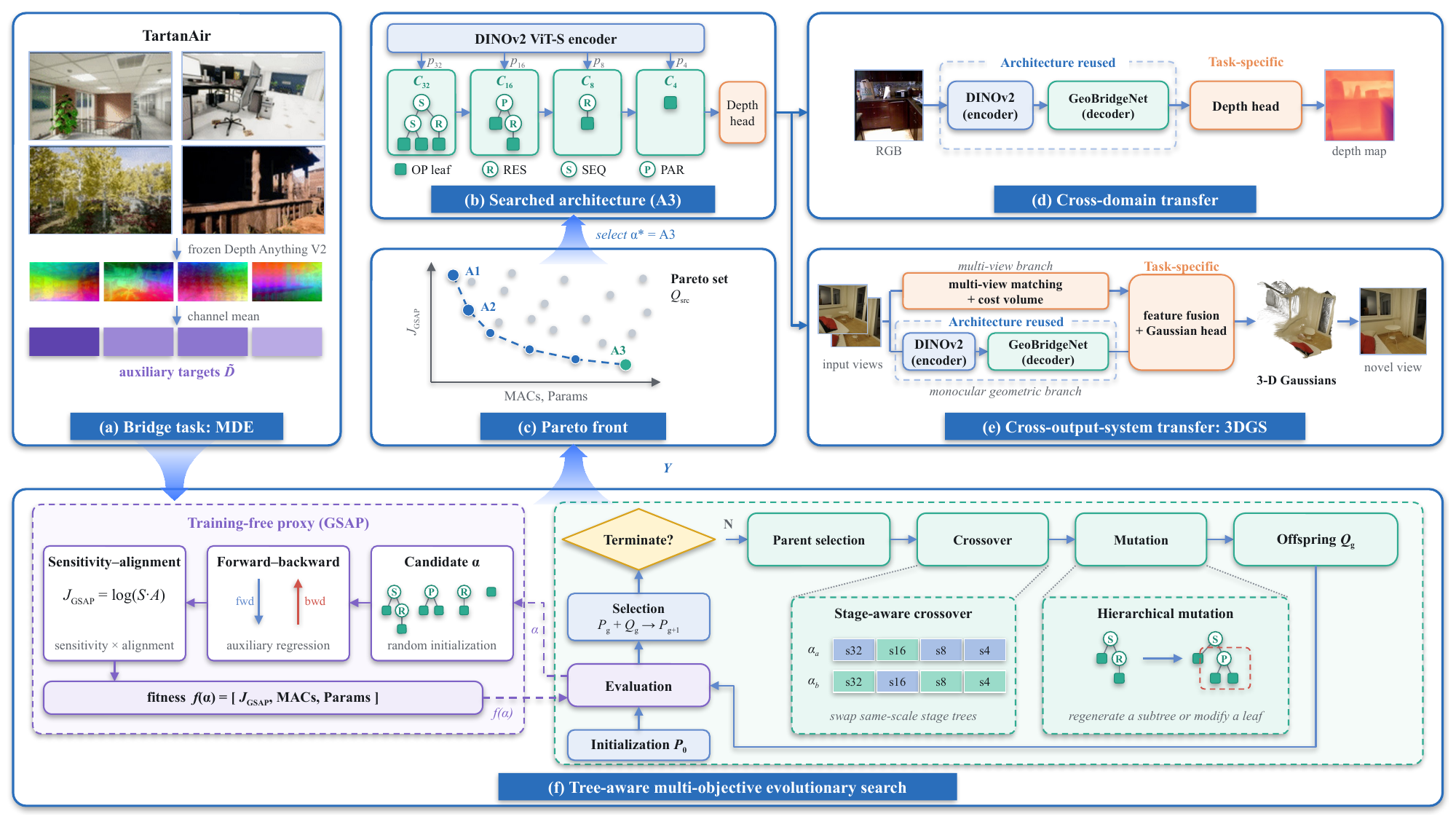}
\caption{\edit{GeoBridge searches once and instantiates the selected decoder topology in target systems. (a) TartanAir images and cached teacher-derived targets define the source evaluation environment. (f) NSGA-II varies stage compositions and primitive configurations through stage-aware crossover and hierarchical mutation; GSAP, MACs, and parameter count define candidate fitness. (c) The Pareto set supplies (b) the selected GeoBridgeNet-A3 structure. Its topology is retained for (d) depth estimation on real images and (e) the monocular branch of a Gaussian reconstruction system. Each target system trains its own weights and prediction interface.}}
\label{fig:generality}
\end{figure*}
}

\newcommand{\FigureOverall}{%
\begin{figure*}[t]
\centering
\includegraphics[width=\textwidth]{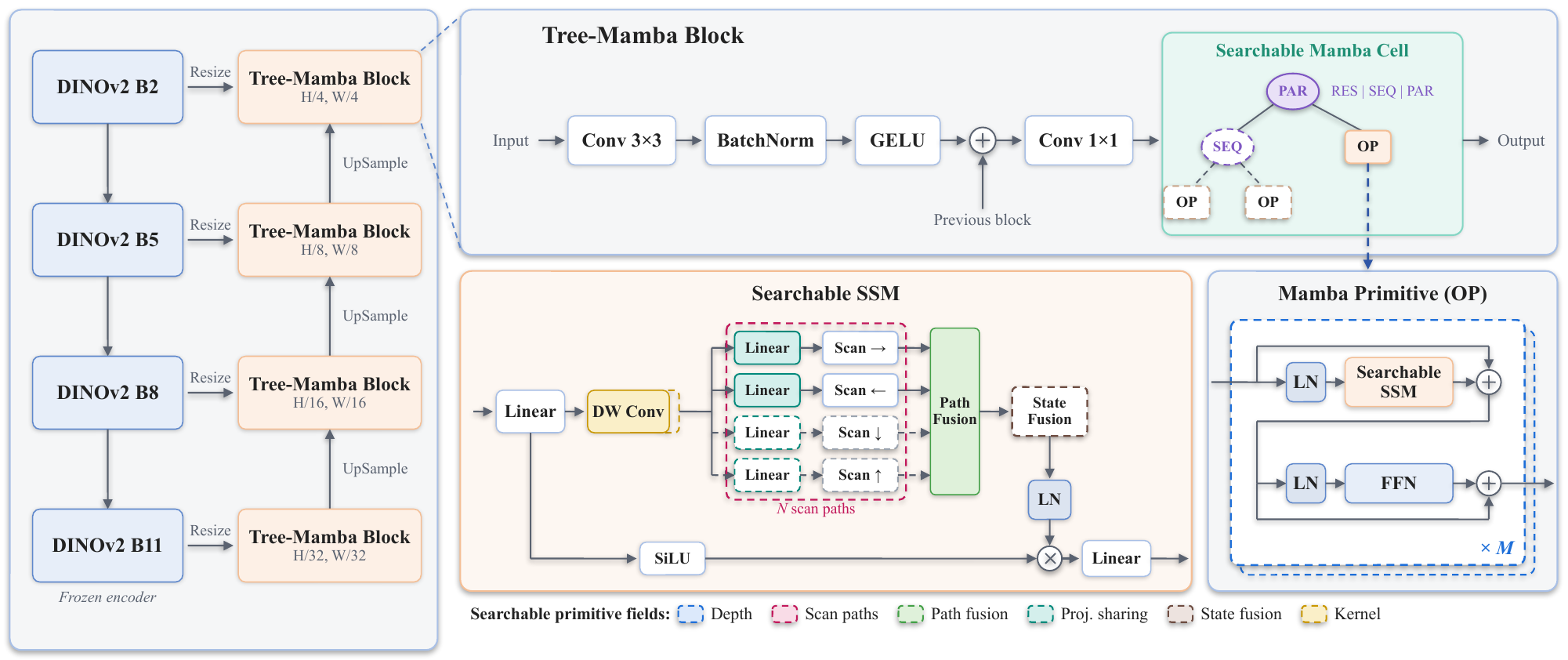}
\caption{\edit{Fixed scaffold and searchable stage computation in GeoBridge. DINOv2 ViT-S blocks B2, B5, B8, and B11 supply features resized to 1/4--1/32 of the input resolution. Lateral fusion and upsampling define the connections between four Tree-Mamba stages. Inside each stage, a functional tree determines the composition of Mamba primitives. Search varies this internal topology and its primitive configurations while retaining the scaffold. Selected structures are trained with the prediction interface required by each target system.}}
\label{fig:overall}
\end{figure*}
}

\newcommand{\FigureSearchSpace}{%
\begin{figure*}[t]
\centering
\includegraphics[width=\textwidth]{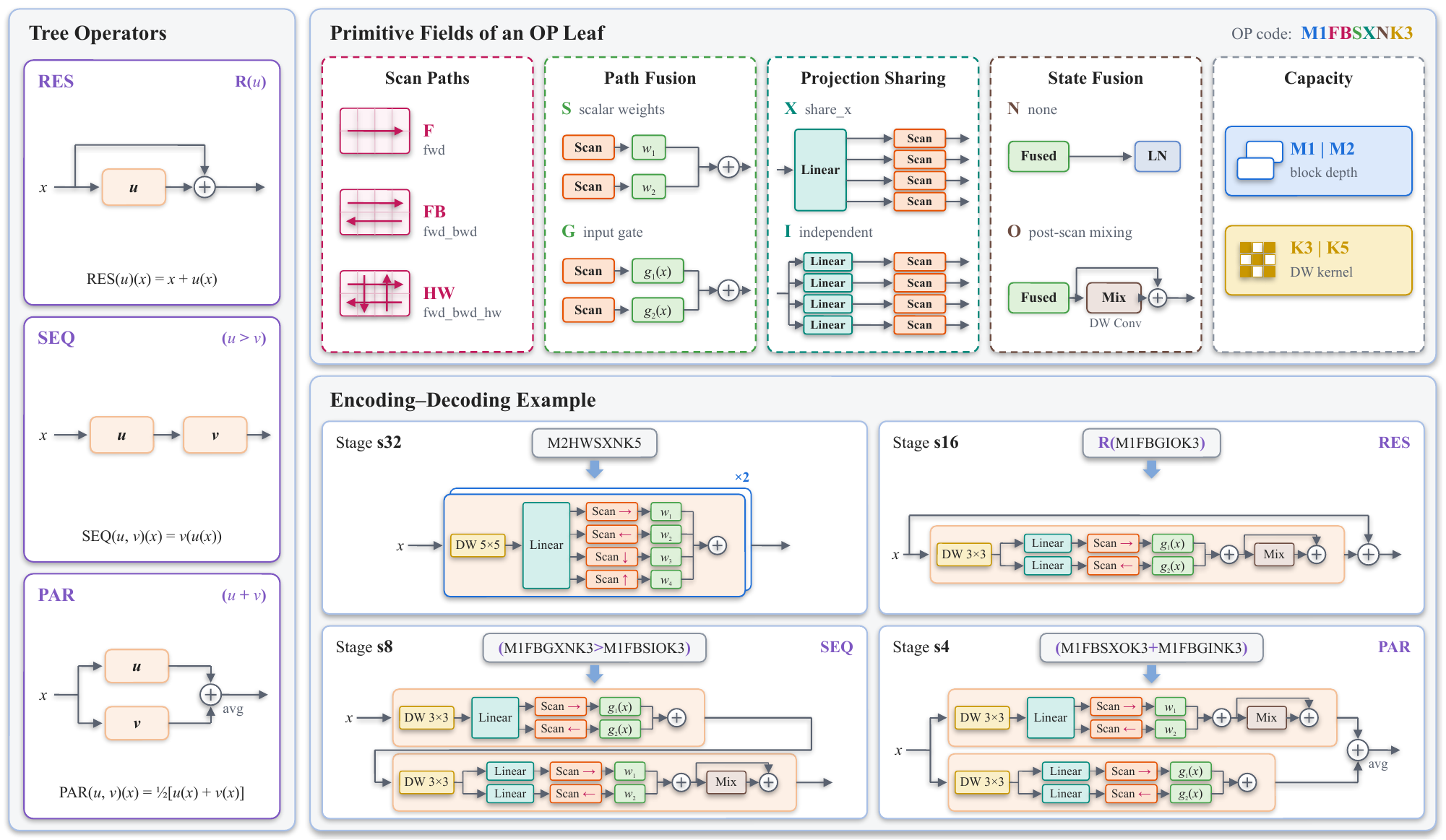}
\caption{\edit{Compositional decoder topology and its tree encoding. Left: RES, SEQ, and PAR define residual paths, serial processing, and parallel aggregation. Top right: the fields of an OP leaf specify its Mamba primitive. Bottom right: an example genotype is decoded into stage computation through recursive composition. Every subtree preserves its stage interface. Table~\ref{tab:search_space} gives the primitive domains and scale-dependent bounds; Fig.~\ref{fig:genotypes} shows the evaluated structures.}}
\label{fig:search_space}
\end{figure*}
}

\newcommand{\FigureProxy}{%
\begin{figure}[t]
\centering
\includegraphics[width=\columnwidth]{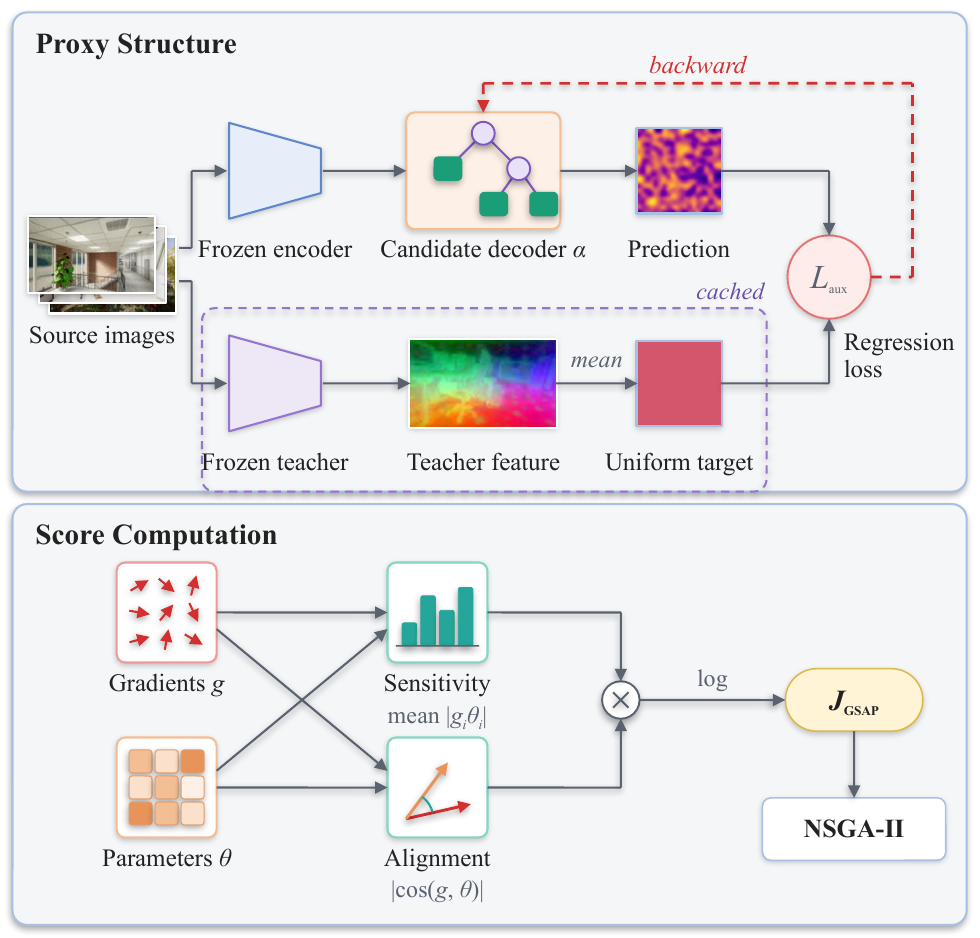}
\caption{{Structure and computation of \gsp{}. Top: a fixed batch of source images passes through the frozen DINOv2 encoder and a randomly initialized candidate decoder, while the frozen Depth Anything V2 teacher provides a spatially uniform auxiliary target that is computed once and cached. One backward pass yields one raw sensitivity--alignment response without updating any parameter. Bottom: the illustrated sensitivity and alignment constitute one raw response; during search, this computation is repeated for three independent decoder initializations, the three raw products are averaged, and the logarithm is then applied to obtain $J_{\mathrm{GSAP}}$, which forms the fitness together with MACs and parameters.}}
\label{fig:gsp}
\end{figure}
}

\newcommand{\FigureSearch}{%
\begin{figure}[t]
\centering
\includegraphics[width=\columnwidth]{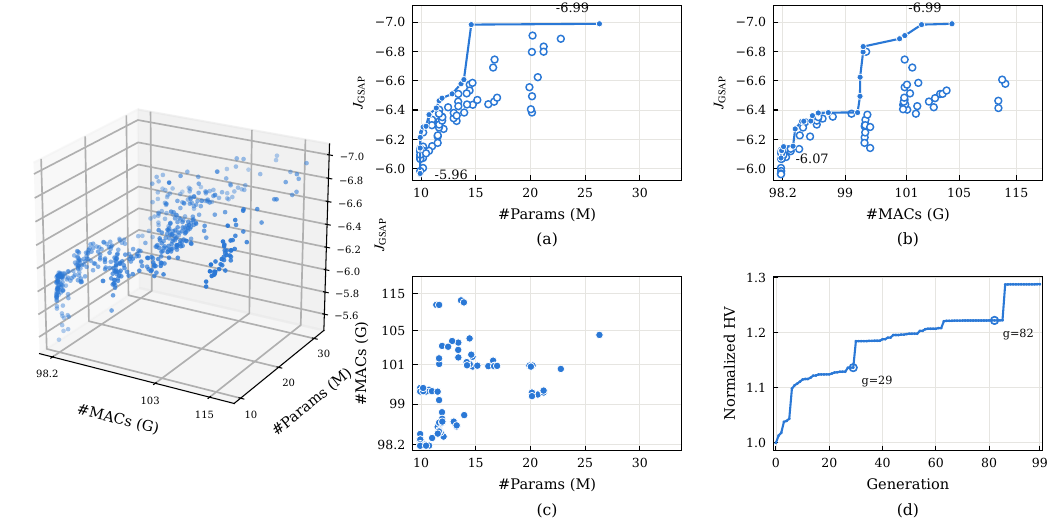}
\caption{{Search dynamics of the 100-generation main run using three-repeat GSAP aggregation. Left: distribution of searched architectures in the three-objective space of $J_{\mathrm{GSAP}}$, MACs, and parameters. (a)--(c) Pairwise projections of the same objective space, with the non-dominated boundary highlighted. (d) Normalized hypervolume over generations; generations 29 and 82 are marked to emphasize that improvement continues beyond the early search before the late-stage plateau. This run is separate from the encoding comparison in Table~\ref{tab:encoding_proxy}.}}
\label{fig:search}
\end{figure}
}

\newcommand{\FigureGenotypes}{%
\begin{figure*}[t]
\centering
\includegraphics[width=0.8\textwidth]{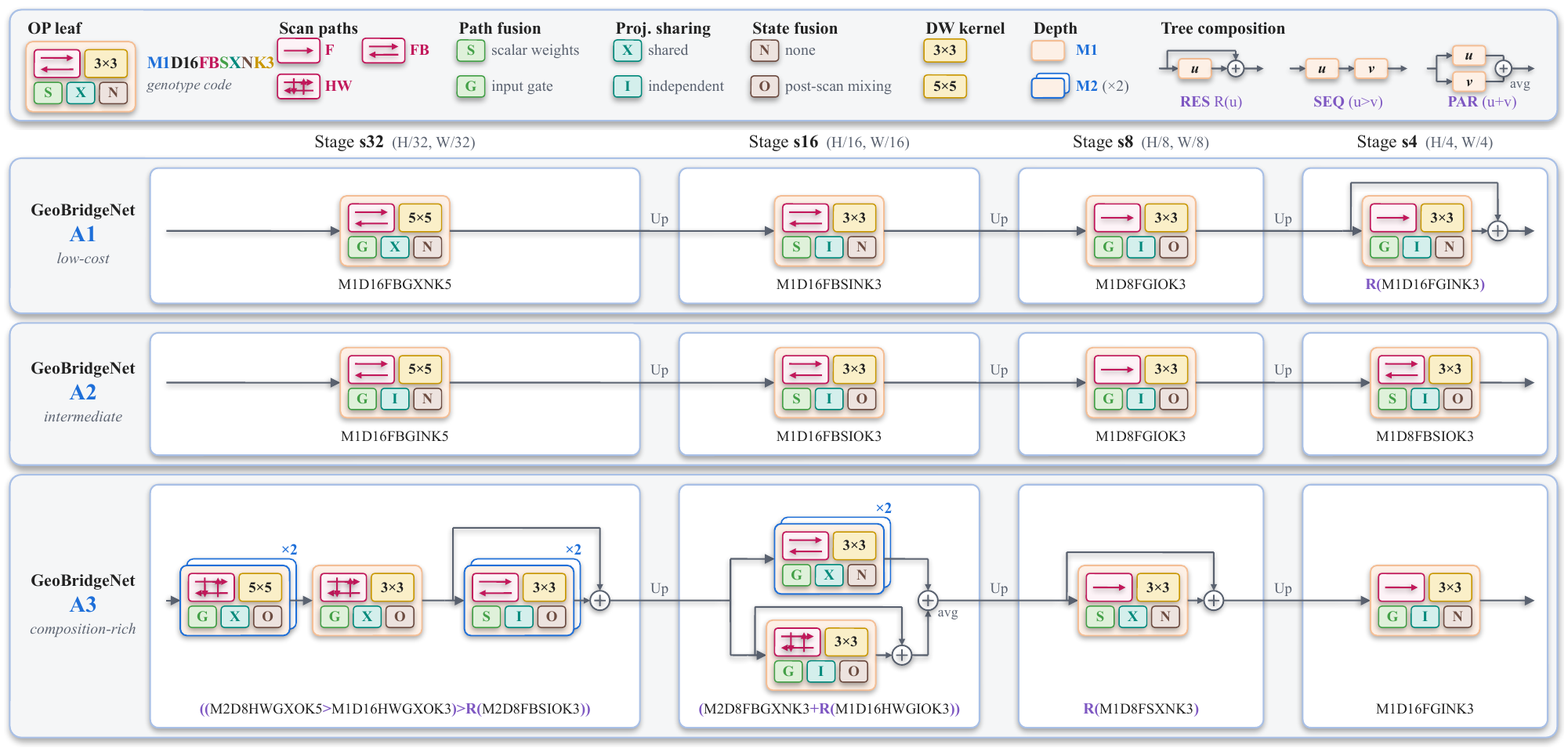}
\caption{Stage-wise architectures of GeoBridgeNet-A1, A2, and A3, decoded from their genotypes. Each row shows the searched Mamba cells of one variant in decoding order from $s32$ to $s4$; Up denotes the fixed lateral fusion and upsampling of the scaffold in Fig.~\ref{fig:overall}. Each card is one OP leaf, whose scan paths, depth-wise kernel, path fusion, projection sharing, and state fusion follow the legend, and a stacked card denotes a two-block primitive. RES, SEQ, and PAR compositions are drawn as in Fig.~\ref{fig:search_space}, and the genotype string of each cell is given beneath it.}
\label{fig:genotypes}
\end{figure*}
}

\newcommand{\FigureNyuQualitative}{%
\begin{figure*}[t]
\centering
\includegraphics[width=0.8\textwidth]{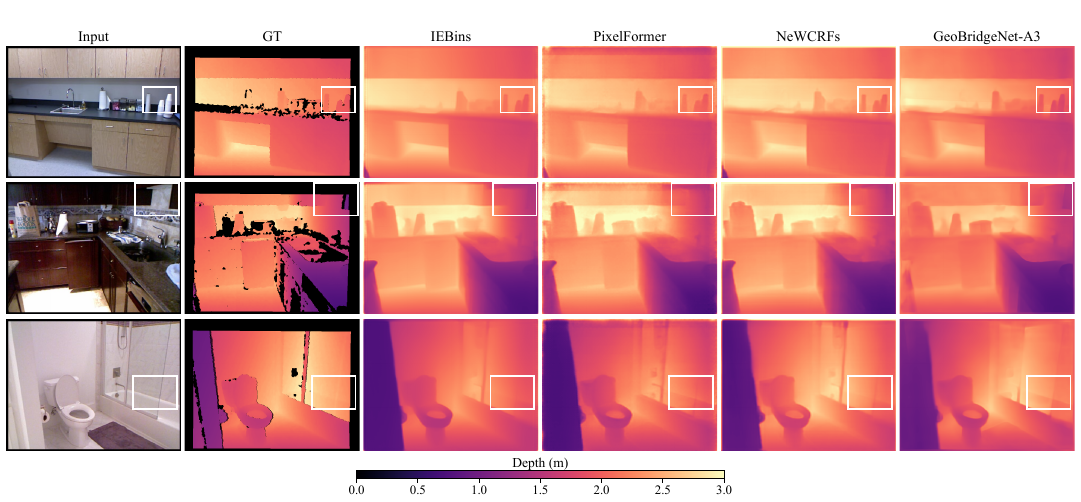}
\caption{NYU Depth V2 qualitative comparison. White boxes mark boundary differences within 0--3 m.}
\label{fig:nyu_sota_qualitative}
\end{figure*}
}

\newcommand{\FigureKittiQualitative}{%
\begin{figure*}[t]
\centering
\includegraphics[width=0.8\textwidth]{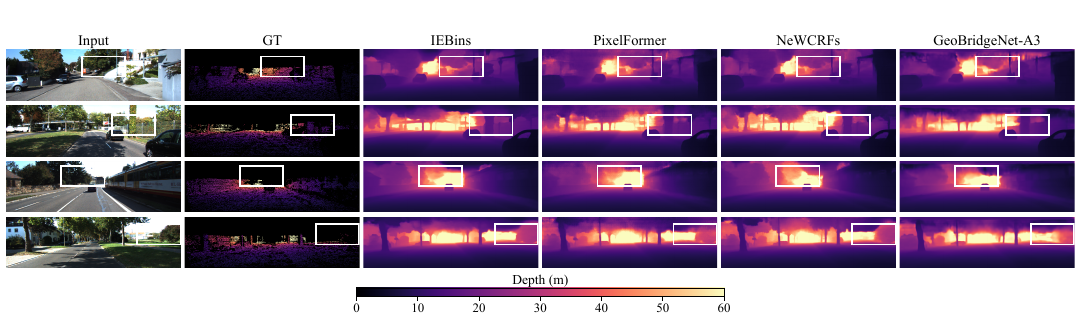}
\caption{KITTI qualitative comparison. White boxes mark boundary differences within 0--60 m.}
\label{fig:kitti_sota_qualitative}
\end{figure*}
}

\newcommand{\FigureReTenKQual}{%
\begin{figure*}[t]
\centering
\includegraphics[width=0.8\textwidth]{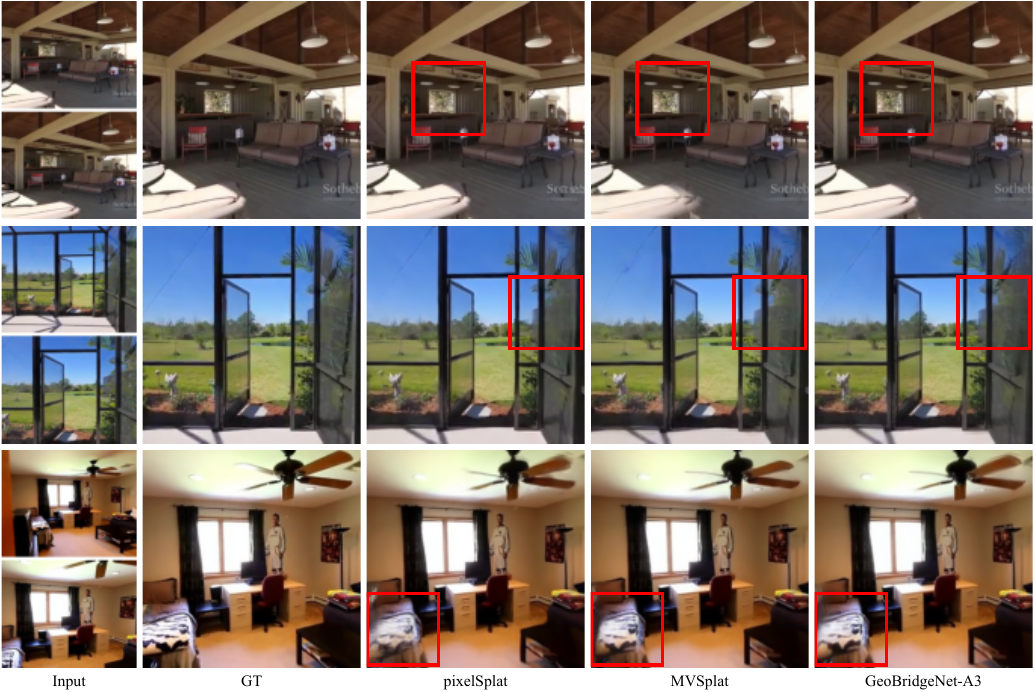}
\caption{RealEstate10K novel-view comparison for GeoBridgeNet-A3. Red boxes mark geometric and detail differences.}
\label{fig:re10k_qualitative}
\end{figure*}
}

\newcommand{\FigureCosThreeDQualitative}{%
\begin{figure*}[t]
\centering
\includegraphics[width=0.8\textwidth]{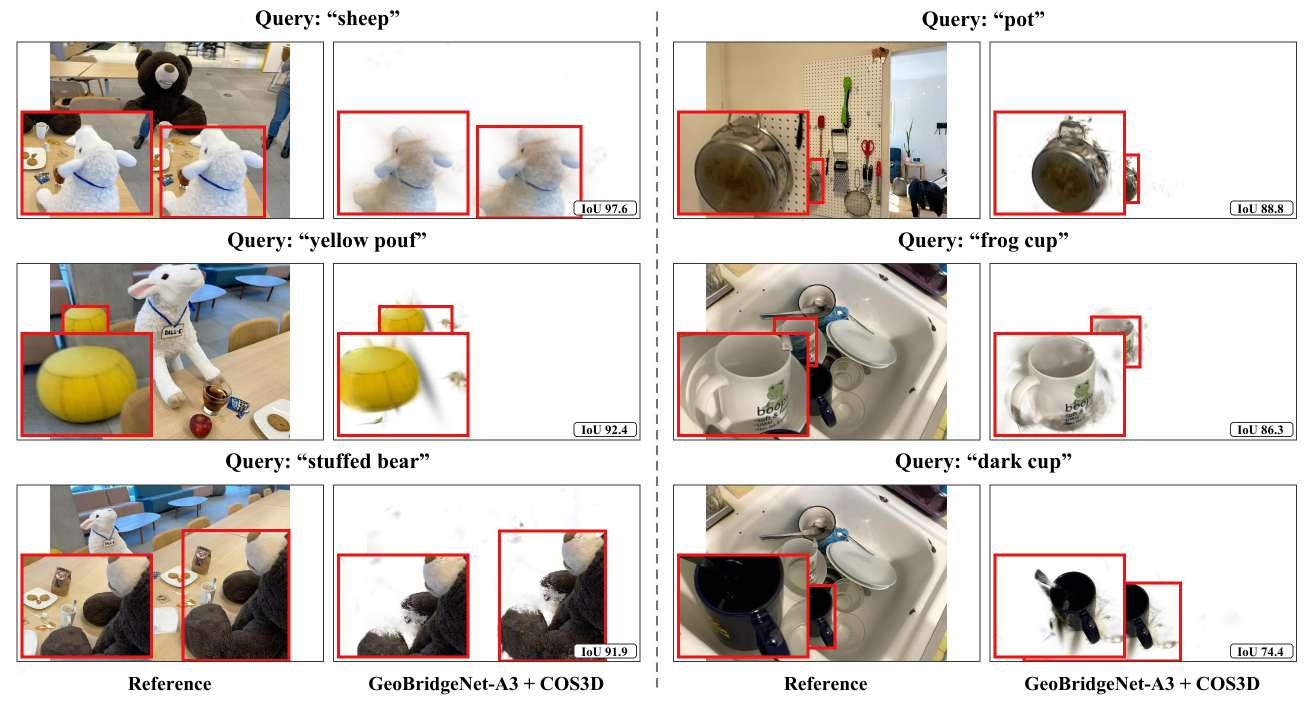}
\caption{Open-vocabulary 3-D segmentation and object isolation with the GeoBridgeNet-A3-initialized Gaussian model. Red boxes mark queried objects, and labels report IoU (\%).}
\label{fig:cos3d_qualitative}
\end{figure*}
}